\def\year{2022}\relax
\documentclass[letterpaper]{article} 
\usepackage{aaai22}  
\usepackage{times}  
\usepackage{helvet}  
\usepackage{courier}  
\usepackage[hyphens]{url}  
\usepackage{graphicx} 
\usepackage{natbib}  
\usepackage{caption} 
\DeclareCaptionStyle{ruled}{labelfont=normalfont,labelsep=colon,strut=off} 
\usepackage{algorithm}
\usepackage{algorithmic}

\usepackage{newfloat}
\usepackage{listings}
\floatstyle{ruled}
\newfloat{listing}{tb}{lst}{}
\floatname{listing}{Listing}

\usepackage{amsfonts,amsmath,booktabs}

\newtheorem{definition}{Definition}
\newtheorem{lemma}{Lemma}
\newtheorem{theorem}{Theorem}

\title{DuMLP-Pin: A Dual-MLP-Dot-Product Permutation-Invariant Network for Set Feature Extraction}
\author {
    Jiajun Fei,\textsuperscript{\rm 1}
    Ziyu Zhu,\textsuperscript{\rm 1}
    Wenlei Liu,\textsuperscript{\rm 1}
    Zhidong Deng,\textsuperscript{\rm 1}
    Mingyang Li,\textsuperscript{\rm 2}
    Huanjun Deng,\textsuperscript{\rm 2}
    Shuo Zhang \textsuperscript{\rm 2}
}
\affiliations {
    \textsuperscript{\rm 1} Institute for Artificial Intelligence at Tsinghua University (THUAI),
    State Key Laboratory of Intelligent Technology and Systems,
    Beijing National Research Center for Information Science and Technology (BNRist),
    Department of Computer Science and Technology, Tsinghua University, Beijing 100084, China\\
    \textsuperscript{\rm 2} Alibaba Group, Hangzhou 310052, China\\
    feijj20@mails.tsinghua.edu.cn, zhuziyu.edward@gmail.com, \{liuwl2020, michael\}@mail.tsinghua.edu.cn, \{mingyangli, huanjun.dhj\}@alibaba-inc.com, ygzhangsoya@qq.com
}

\begin{document}

\maketitle

\begin{abstract}
    Existing permutation-invariant methods can be divided into two categories according to the aggregation scope, i.e. global aggregation and local one. Although the global aggregation methods, e. g., PointNet and Deep Sets, get involved in simpler structures, their performance is poorer than the local aggregation ones like PointNet++ and Point Transformer. It remains an open problem whether there exists a global aggregation method with a simple structure, competitive performance, and even much fewer parameters. In this paper, we propose a novel global aggregation permutation-invariant network based on dual MLP dot-product, called DuMLP-Pin, which is capable of being employed to extract features for set inputs, including unordered or unstructured pixel, attribute, and point cloud data sets. We strictly prove that any permutation-invariant function implemented by DuMLP-Pin can be decomposed into two or more permutation-equivariant ones in a dot-product way as the cardinality of the given input set is greater than a threshold. We also show that the DuMLP-Pin can be viewed as Deep Sets with strong constraints under certain conditions. The performance of DuMLP-Pin is evaluated on several different tasks with diverse data sets. The experimental results demonstrate that our DuMLP-Pin achieves the best results on the two classification problems for pixel sets and attribute sets. On both the point cloud classification and the part segmentation, the accuracy of DuMLP-Pin is very close to the so-far best-performing local aggregation method with only a 1-2\% difference, while the number of required parameters is significantly reduced by more than 85\% in classification and 69\% in segmentation, respectively. The code is publicly available on \url{https://github.com/JaronTHU/DuMLP-Pin}.
\end{abstract}

\section{Introduction}

Convolutional neural networks (CNNs) have superiority in feature extraction on structured data like text, images, or other sequences. For unordered or unstructured set inputs, however, permutation-invariant methods seem to be preferred. 

According to the aggregation scope, existing permutation-invariant networks, hereafter referred to as Pin, can be divided into global aggregation and local aggregation methods. Pioneer work uses global aggregation to achieve permutation-invariance. For instance, PointNet \cite{qi2017pointnet} employs global max pooling, while global sum pooling is adopted in Deep Sets \cite{zaheer2017deep}. In general, there are simple structures in global aggregation methods. But they are not good at extracting local features. Compared to these, local aggregation methods like PointNet++ \cite{qi2017pointnet++} and Point Transformer \cite{zhao2020point} overcome such limitation by defining the local aggregation operation and scope. Accordingly, they yield better performance, though they usually have more complex structures.

In this paper, we propose a dual MLP dot-product permutation-invariant network (DuMLP-Pin) for set feature extraction. The key to DuMLP-Pin is the use of two multilayer perceptrons (MLPs) in order to aggregate information globally. It does gain permutation-invariance through the dot-product of outputs of two MLPs. We strictly prove that the dot-product decomposition always exists for any permutation-invariant function. Compared to other permutation-invariant methods on the same task, DuMLP-Pin is highly parameter-efficient, which means that it only needs a small number of parameters. Furthermore, we also show that the optimization of DuMLP-Pin can be viewed as the constrained optimization of Deep Sets under certain conditions.

The contributions of this paper are summarized as follows:
\begin{itemize}
  \item We propose a novel global aggregation permutation-invariant network based on dual MLP dot-product decomposition, called DuMLP-Pin.
  \item We strictly prove that a dot-product decomposition for any permutation-invariant function implemented by DuMLP-Pin always exists if and only if the cardinality of input sets is greater than a pre-specific threshold.  
  \item We explain the high efficiency of DuMLP-Pin and show that under certain conditions DuMLP-Pin can be considered as Deep Sets with strong constraints.
\end{itemize}

We evaluate DuMLP-Pin on several problems with different types of data elements. Experimental results demonstrate that DuMLP-Pin yields the best performance on both the pixel classification and the attribute set one, compared to other permutation-invariant methods. On the point cloud classification, the performance of the proposed method almost reaches the best-performing local aggregation method with a difference of only 1.4\%, but the number of required parameters is remarkably reduced by more than 85\%. In addition, it is also approximately close to the best local aggregation method on the point cloud part segmentation task in performance, where there is only a 1.8\% difference between them, but the reduction in the number of required parameters exceeds 69\%. Technical appendix is available on \url{https://arxiv.org/abs/2203.04007}.

\section{Related Work}

\subsection{Global Aggregation}

Recently, a collection of researches focus on constructing feature extractors for unordered or unstructured data inputs. In pioneer work, global pooling functions are often employed to gain permutation-invariance. Specifically, PointNet \cite{qi2017pointnet} utilizes max pooling to directly tackle unordered point cloud data. Feature transformations are also exploited to boost performance in such network architecture. Although PointNet mainly gets involved in point cloud problems, it is easily transferred to other unordered data as well. Meanwhile, Deep Sets \cite{zaheer2017deep, ravanbakhsh2016deep} makes use of sum pooling to aggregate information. It also points out that conditional mapping could be adopted as additional meta-information is available. The core of Deep Sets is to add up all representations and then apply nonlinear transformations. BRUNO \cite{korshunova2018bruno} is a recurrent model for Bayesian inference on set inputs, where joint distribution over observations is permutation-invariant. Besides the above-mentioned simple pooling functions, attention-based architecture is also popular when building permutation-invariant or permutation-equivariant functions. \citeauthor{ilse2018attention} proposes an attention-based aggregation operator to learn the Bernoulli distribution related to multiple instance learning problems \cite{maron1998framework}. Set Transformer \cite{lee2019set} is another Transformer-based neural network that is designed to process sets of data. In such a model, several attention-related set operations are presented under different problem settings. There are many research works associated with permutation-invariance and permutation-equivariance in Transformer \cite{zhao2020exploring,khan2021transformers}, where some of them leverage permutation-invariance, but some do not. Unlike previous methods directly approximating global features, RepSet \cite{skianis2020rep} generates them by solving the maximum matching problem between the elements of two sets. 

Global aggregation methods have simple structures, whose permutation-invariance is gained by either global pooling functions or attention mechanisms. Such two types of methods, however, have different limitations. \citeauthor{wagstaff2019limitations} analyzes the limitation of pooling architectures and provides a solution by imposing a constraint on the dimensionality of the latent space. For Transformer-based global aggregations, there is usually a relatively heavy computational complexity. Some researchers argue that full attention in Transformer is not necessary \cite{sukhbaatar2019adaptive,kitaev2019reformer,liu2021swin}, and accordingly, different ways are delivered to do computation more efficiently.

\subsection{Local Aggregation}

Global aggregation methods are always weak in capturing local information, which generally results in relatively poor performance, especially in the point cloud settings. To overcome this drawback, local aggregation methods are proposed. PointNet++ \cite{qi2017pointnet++} captures local structures by recursively applying PointNet on the partitioning of point cloud inputs. In the PointNet++ model, two density adaptive layers, i.e. multi-scale grouping and multi-resolution grouping, are proposed to group local regions and combine features from different scales. PointASNL \cite{yan2020pointasnl} focuses on tackling point clouds with noise and presents adaptive sampling and local-nonlocal modules to seize local and long-range dependencies. Point Cloud Transformer \cite{guo2020pct} is also based on Transformer architectures. Unlike Set Transformer, however, it uses the offset-attention module with neighbor embedding for augmented local feature aggregation. Point Transformer \cite{zhao2020point} establishes attention mechanisms based on \cite{zhao2020exploring}. It also employs relative positional encoding to aggregate local information. In summary, local aggregation methods generally have more complex structures since they have to carefully define aggregation operations and scopes.

\section{DuMLP-Pin}

In this section, we first formulate the permutation-invariant learning problem and then propose a novel global aggregation permutation-invariant network based on dual MLP dot-product decomposition (DuMLP-Pin).

\subsection{Problem Formulation}

Suppose that $P$ denotes the permutation matrix and $\mathcal{P}_N$ the set of all permutation matrices with size $N\times N$, where $N$ is the number of elements in a given input set. Let $p$ and $r$ indicate the dimension of the input and output vectors, respectively. According to the definition of permutation-invariance and permutation-equivariance \cite{zaheer2017deep}, we can rewrite them in a matrix form.

\begin{definition}[Permutation-invariant]
  The function $f:\mathbb{R}^{N\times p}\rightarrow\mathcal{Y}$ is permutation-invariant if
  \begin{equation}
    f(PX)=f(X)\ \ \text{ for }\ \forall X\in\mathbb{R}^{N\times p},\ \forall P\in\mathcal{P}_N.
  \end{equation}
\end{definition}
\begin{definition}[Permutation-equivariant]
  The function $g:\mathbb{R}^{N\times p}\rightarrow\mathbb{R}^{N\times r}$ is permutation-equivariant if
  \begin{equation}
    g(PX)=Pg(X)\ \ \text{ for }\ \forall X\in\mathbb{R}^{N\times p},\ \forall P\in\mathcal{P}_N.
  \end{equation}
\end{definition}

In this paper, we pay attention to the set classification problem with supervised learning. Accordingly, the permutation-invariant learning problem can be described as follows:
\begin{definition}[Permutation-invariant Learning]
  Given M training samples in the form of $\left\{(X_i,y_i)\right\}_{i=1}^M$, where $X_i\in\mathbb{R}^{N\times p}$ represents a set input and $y_i\in\mathcal{Y}$ a label, find the function $f:\mathbb{R}^{N\times p}\rightarrow\mathcal{Y}$ such that the corresponding $f$-related performance metric evaluated on test sets is maximized.
\end{definition}

\subsection{Minimum Dimension Decomposition}

We can construct a bijection $u:\mathbb{R}^{m\times n}\rightarrow\mathbb{R}^{mn}$ through simply reshaping, which means that $u(x)_{in+j}=x_{i,j}$. Clearly, if a function can represent any matrices of size $m\times n$, it can then express any vectors of length $mn$ by adding a flattening module. In the following discussion, we will focus on the representation of feature matrices instead of feature vectors. First of all, we introduce Lemma \ref{lem:1} for minimum dimension decomposition (MDD).
\begin{lemma}[MDD]
  $\forall A\in\mathbb{R}^{m\times n}, m\leq n, \forall l\geq m, \forall B$ $\in\{X\in\mathbb{R}^{m\times l}\ \big|\ \mbox{rank}(X)=m\}$. Let $S=\left\{C\in\mathbb{R}^{l\times n}|\right.$ $\left.BC=A\right\}$, then
  \begin{equation}
    S\neq \Phi\text{ and }S=\left\{C_p+X_h\Lambda|\Lambda\in\mathbb{R}^{(l-m)\times n}\right\},
  \end{equation}
  where $C_p$ stands for a particular solution of $BC=A$, and the columns of $X_h$ are basis of Ker$(B)$.
  \label{lem:1}
\end{lemma}

The proof is supplemented in the technical appendix. Obviously there exist infinite tuples $(B, C)$ for any $A$ through MDD. Therefore, we have more choices when we represent $A$ through the decomposition of $B$ and $C$.

\subsection{Permutation-Invariance through Dot-Product Decomposition}

By substituting matrices into matrix functions in Lemma \ref{lem:1}, we may always decompose one matrix function by two matrix functions. In Theorem \ref{the:1}, we show that any permutation-invariant function can be decomposed into two permutation-equivariant functions through a dot-product decomposition.

\begin{theorem}
  \label{the:1}
  For any permutation-invariant function $f:\mathbb{R}^{N\times p}\rightarrow\mathbb{R}^{s\times t}$, there exist two permutation-equivariant functions $g^{(1)}:\mathbb{R}^{N\times p}\rightarrow\mathbb{R}^{N\times s}, g^{(2)}:\mathbb{R}^{N\times p}\rightarrow\mathbb{R}^{N\times t}$, such that
  \begin{equation}
    \forall X\in\mathbb{R}^{N\times p},f(X)=\left(g^{(1)}(X)\right)^Tg^{(2)}(X)
  \end{equation}
  \begin{equation*}
    \text{if and only if } N\geq\min\{s,t\}.
  \end{equation*}
\end{theorem}

The proof is also seen in the technical appendix for details. Unlike the theorems given in previous work \cite{qi2017pointnet,zaheer2017deep}, Theorem \ref{the:1} discuss the relationship between the shape of the feature matrix $\min\{s,t\}$ and the cardinality of the input set $N$. An intuition is that when $N<\min\{s,t\}$, the term $\left(g^{(1)}(X)\right)^Tg^{(2)}(X)$ cannot be full-rank according to linear algebra. Therefore, it cannot represent full-rank feature matrices, which extremely limits the expression capability. In practice, it is much easier to construct a permutation-equivariant function since any function that processes each element independently is permutation-equivariant. It is also the main reason that we refer to two permutation-equivariant functions to represent one permutation-invariant function. When generating feature vectors of length $C$, the outputs of both MLPs can be only as small as $\sqrt{C}$ (when $N\geq\sqrt{C}$), which leads to a great reduction of parameters.

\subsection{Aggregation Block and Broadcast Block}

\begin{figure}[htbp]
  \centering
  \includegraphics[width=0.7\linewidth]{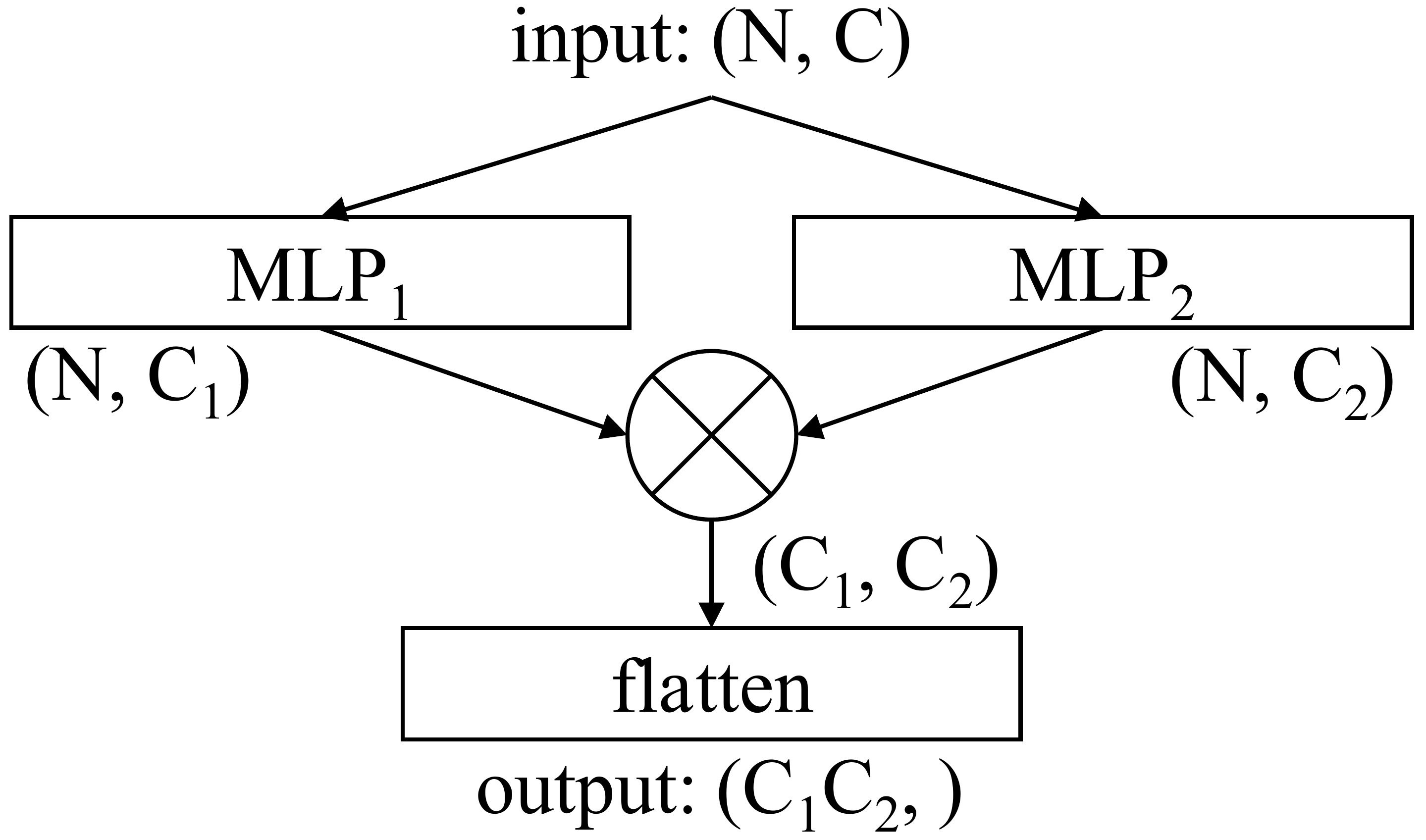}
  \centerline{(a) Aggregation block}
  \includegraphics[width=0.7\linewidth]{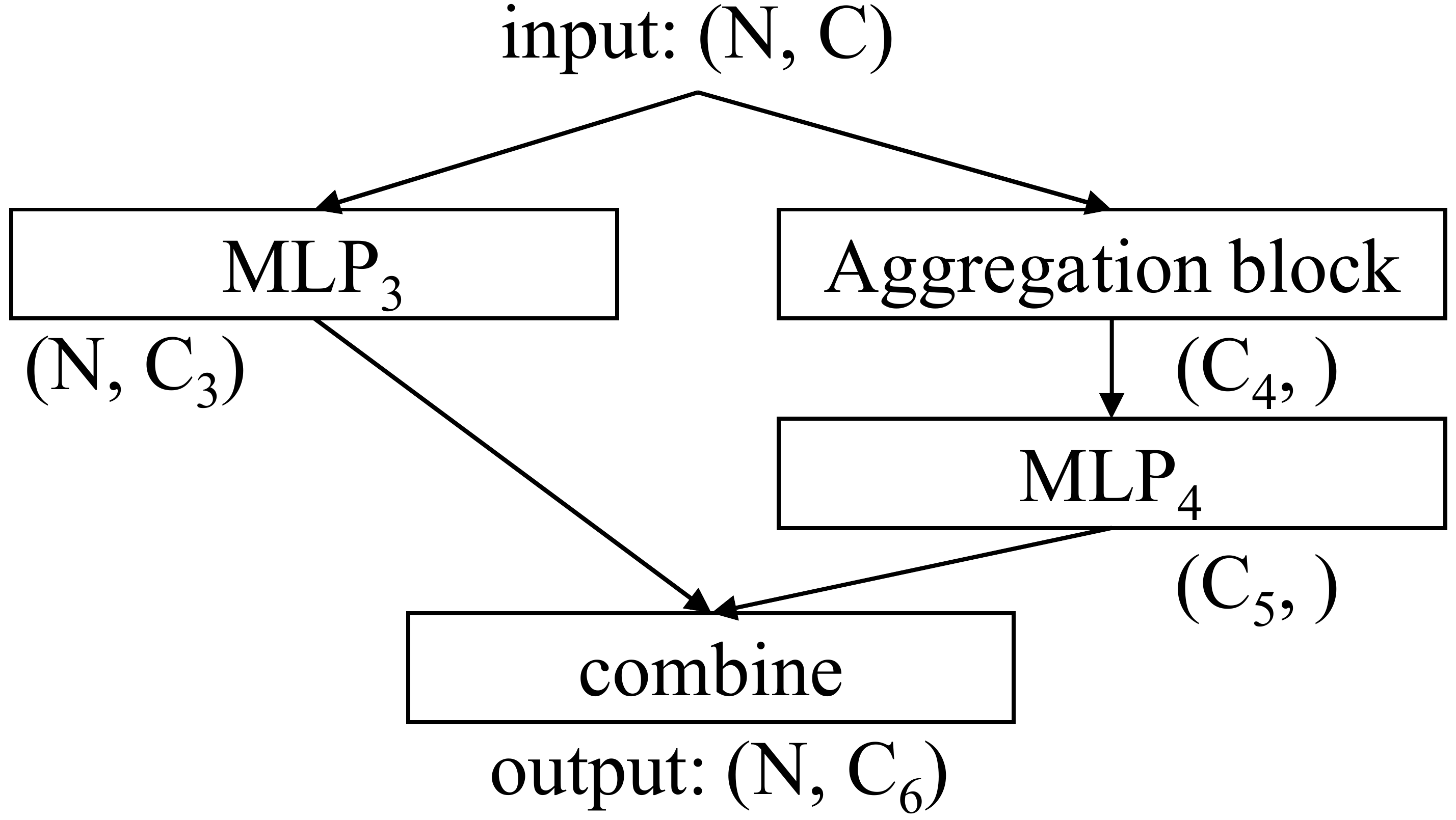}
  \centerline{(b) Broadcast block}
  \caption{Structure design for aggregation block and broadcast block.}
  \label{fig:1}
\end{figure}

DuMLP-Pin is mainly constructed by two kinds of blocks: aggregation block and broadcast block. The aggregation block, which performs global feature aggregation, consists of two vanilla MLPs. It achieves permutation-invariance through the dot-product of the outputs of these MLPs. However, sometimes we need not only set features that represent all unordered or unstructured elements, but also element features for element label prediction. Thus we need to broadcast the set feature to each element. In the following, we design the broadcast block to address such a problem. Formally, let element feature $x\in\mathbb{R}^{d_x}$, set feature $y\in\mathbb{R}^{d_y}$. Then we use Equation \ref{equ:6} to generate combined feature $z\in\mathbb{R}^{d_z}$.
\begin{equation}
  z=\left(W_{xy}\cdot x\right)\cdot y+W_{x}\cdot x+W_{y}\cdot y+b,
  \label{equ:6}
\end{equation}
where $W_{xy}\in\mathbb{R}^{d_z\times d_y\times d_x}, W_x\in\mathbb{R}^{d_z\times d_x}, W_y\in\mathbb{R}^{d_z\times d_y},$ and $b\in\mathbb{R}^{d_z}$. Although $W_{xy}$ involves direct interaction between $x$ and $y$, it increases computational burden tremendously. We observe that involving such term only leads to insignificant performance improvement in experiment. Consequently, we balance efficiency and effectiveness by setting $W_{xy}=0$. The structures of aggregation block and broadcast block are shown in Figure \ref{fig:1} (a) and (b), respectively.

\subsection{Nonlinear Activation Function}

Nonlinear activation functions are essential for the success of deep learning. Besides the common-used ReLU \cite{nair2010rectified}, there are many works focusing on designing novel nonlinear activation functions, including ELU \cite{clevert2015fast}, SELU \cite{klambauer2017self}, Mish \cite{misra2019mish}, etc. We do not discuss general preferences on the choice of nonlinear activation functions. Instead, we only focus on the last layers of two MLPs in the aggregation block.

Without loss of generality, suppose that both MLPs comprise only one linear layer without any bias. Let $W_1$ and $W_2$ be weights. $\alpha_1$ and $\alpha_2$ are assumed to be nonlinear activation functions used by $\text{MLP}_1$ and $\text{MLP}_2$, respectively. Since we argue both MLPs must be permutation-equivariant, here we only discuss element-wise activation functions like ReLU and set-wise ones like softmax. Then for input $X\in\mathbb{R}^{N\times C}$, the output is computed as:
\begin{equation}
  f(x)=\alpha_1^\prime(W_1^TX^T)\alpha_2(XW_2),
  \label{equ:12}
\end{equation}
where $\alpha_1^\prime$ satisfies that $\alpha_1^\prime(W_1^TX^T)=\alpha_1^T(XW_1)$. We find that the model always shows better performance when using at least one nonlinear activation function. Why nonlinear activation functions are essential? Assume that both MLPs do not use nonlinear activation function. In this case, Equation \ref{equ:12} turns into Equation \ref{equ:2}, i.e.,
\begin{equation}
  f(x)=W_1^TX^TXW_2=W_1^T(X^TX)W_2.
  \label{equ:2}
\end{equation}

It is easy to see that $f(X)$ is dominated by $X^TX\in\mathbb{R}^{C\times C}$. This is inappropriate and even destructive especially when $C\ll N$. For example, the feature dimension of the input element is relatively small under some point cloud settings, which is much smaller than the number of points. If no nonlinear activation function is used, then unfeasible constraints that for $X_1,X_2\in\mathbb{R}^{N\times C}, X_1^TX_1=X_2^TX_2\Rightarrow f(X_1)=f(X_2)$ will be imposed. However, using even one nonlinear activation function can overcome this problem by cutting off the direct combination between $X^T$ and $X$.

Besides, we observe that set-wise softmax generally outperforms element-wise activation functions. We argue that set-wise softmax involves direct interaction among elements in inputs, thus leading to better results than element-wise ones. We will discuss this more in the ablation study.

\subsection{Further Discussion}

In this part, we methodologically make comparisons of our DuMLP-Pin with other global aggregation methods including Deep Sets, PointNet, and Set Transformer. 

\textbf{Relation to both Deep Sets \cite{zaheer2017deep} and PointNet \cite{qi2017pointnet}}. Both Deep Sets and PointNet build permutation-invariant functions in a similar manner:
\begin{align*}
  \text{Deep Sets: }f(X)&=\gamma\left(\sum_{x_i}\left\{h(x_i)\right\}\right),\\
  \text{PointNet: }f(X)&=\gamma\left(\max_{x_i}\left\{h(x_i)\right\}\right).
\end{align*}

The main difference lies in the pooling function, which is also the core of permutation-invariance. To illustrate how DuMLP-Pin works, let us suppose that each row is processed independently. It implies that any interaction between elements does not get involved. Thus we make the derivation below:
\begin{align*}
  f(X)=&\left(g^{(1)}(X)\right)^Tg^{(2)}(X)\\
  =&\sum_{x_i}\left(g^{(1)}(x_i)\right)^T\left(g^{(2)}(x_i)\right)\\
  =&\sum_{x_i}h^\prime (x_i).
\end{align*}

Therefore, DuMLP-Pin also takes advantage of the sum pooling over element features. However, unlike $h(x_i)$ in Deep Sets, $h^\prime (x_i)\in\mathbb{R}^{C_1\times C_2}$ is subject to the following strong constraints:
\begin{equation}
  \text{rank}\left[h^\prime (x_i)\right]\leq 1.
\end{equation}

From this perspective, DuMLP-Pin can be regarded as Deep Sets with strong constraints on element features. We argue that the direct optimization of $h$ is generally tougher and more unstable than optimizing $g^{(1)}$ and $g^{(2)}$ concurrently. This is because we shrink the search space by imposing constraints without loss of theoretical expression capability. If we use set-wise softmax, then $g(x_i)$ can be modified as $g(x_i)p(X)$ where $p(X)$ is a global pooling. But it does not change the low-rank constraints.

\textbf{Relation to Set Transformer \cite{lee2019set}}. Although Pooling by Multihead Attention Block (PMA) in Set Transformer is similar to our aggregation block, they are different in many aspects as shown in Equation \ref{equ:1}. Set Transformer is based on Transformer architecture, which implies that it mainly employs linear mapping for feature transformation, while our DuMLP-Pin uses MLPs for the same sake. In addition, the aggregation block does not leverage the self-attention mechanism, layer normalization, and multi-head architecture. The output of Set Transformer has different meanings on different dimensions (i.e. sample and feature). But our DuMLP-Pin takes both dimensions as features, making it natural to flatten for further use.

\begin{equation}
  \begin{aligned}
    f_\text{PMA}(X)=&\ \text{Attention}\left(S,X,X\right),\\
    f_\text{DuMLP-Pin}(X)=&\ \text{Flatten}\left[\text{MLP}_1\left(X\right)^T\text{MLP}_2\left(X\right)\right].
  \end{aligned}
  \label{equ:1}
\end{equation}

\textbf{Possibility of High-order Dot-Product Decomposition}. DuMLP-Pin can be viewed as Deep Sets with strong constraints on element features, which is derived by raising feature dimensions from 1D vector to 2D matrix. Is it possible that we construct 3D, 4D, or even $n$D feature tensors to further shrink search space and boost performance? We dive into such a problem and formulate the result as Theorem \ref{the:2}.

\begin{theorem}
  \label{the:2}
  For any permutation-invariant function $f:\mathbb{R}^{N\times p}\rightarrow\mathbb{R}^{c_1\times\cdots\times c_n}$, there exist $n$ permutation-equivariant functions $g^{(1)}:\mathbb{R}^{N\times p}\rightarrow\mathbb{R}^{N\times c_1}, \cdots, g^{(n)}:\mathbb{R}^{N\times p}\rightarrow\mathbb{R}^{N\times c_n}$, such that ($1\leq a_j\leq c_j$)
  \begin{equation*}
    \forall X\in\mathbb{R}^{N\times p},f(X)_{a_1,\cdots,a_n}=\sum_{i=1}^N\prod_{j=1}^n\left(g^{(j)}(X)\right)_{i,a_j}
  \end{equation*}
  \begin{equation*}
    \text{if }N\geq\frac{\prod_{j=1}^nc_j}{\max_jc_j}\text{ and only if }N\geq\frac{\prod_{j=1}^nc_j}{\sum_{j=1}^nc_j}
  \end{equation*}
  \begin{equation*}
    \text{and }N\geq\max_{C\subset\left\{c_1, \cdots, c_n\right\}}\min\left\{\prod_{c\in C}c, \prod_{c\in\overline{C}}c \right\}.
  \end{equation*}
\end{theorem}

The proof is also supplemented in the technical appendix. We compare the performance of different feature orders in the ablation study.

\section{Experiments}

On several problems with different types of data elements, we evaluate the performance of our DuMLP-Pin. In the experiments, we employ the MNIST for pixel set classification, the CelebA for attribute set anomaly detection, the ModelNet40 for point cloud classification, and the ShapeNetPart for point cloud part segmentation, respectively. We independently run all experiments 5 times and report the evaluation metric on test sets. More details about the architectures and hyper-parameters are available in the technical appendix.

\subsection{Pixel Set Classification}

The MNIST \cite{lecun1998gradient} is a well-known handwritten digit classification dataset with 60,000 images in the training set and 10,000 images in the test set. We transform images into pixel sets by appending relative coordinates to intensities and shuffling the order randomly. Accordingly, our DuMLP-Pin is compared with PointNet and PointNet++, respectively, since they also consider images as pixel sets. We use the error rate as the evaluation metric. Both results quoted are taken from the cited papers.

\begin{table}[htbp]
  \centering
  \begin{tabular}{ccc}
    \toprule
    Method & \#Params(M) & Error Rate(\%) \\
    \midrule
    PointNet \shortcite{qi2017pointnet} & 3.47 & 0.78\\
    PointNet++ \shortcite{qi2017pointnet++} & 1.47 & 0.51\\
    \midrule
    DuMLP-Pin-S & 0.28 & 0.80 $\pm$ 0.053\\
    DuMLP-Pin-L & 1.13 & \textbf{0.48} $\pm$ 0.034\\
    \bottomrule
  \end{tabular}
  \caption{The MNIST pixel set classification results.}
  \label{tab:1}
\end{table}

\begin{figure}[htbp]
  \centering
  \includegraphics[width=0.7\linewidth]{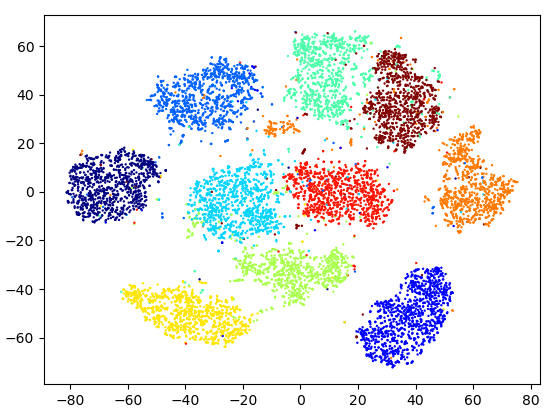}
  \caption{The t-SNE embeddings of DuMLP-Pin-S on the MNIST test set. Feature vectors are generated by one aggregation block with only 17.9K parameters.}
  \label{fig:3}
\end{figure}

We evaluate two models of different sizes: DuMLP-Pin-S with only one aggregation block and DuMLP-Pin-L with one aggregation block and two broadcast blocks. From Table \ref{tab:1}, it is readily observed that DuMLP-Pin-S performs slightly worse than PointNet while DuMLP-Pin-L is slightly better than PointNet++. However, unlike PointNet++ as a local aggregation method, DuMLP-Pin has no access to local structures, making the results more competitive. It is possible to construct a DuMLP-Pin model with better hyper-parameters, which may further improve the performance. The experiment shows that DuMLP-Pin has the potential to process simple images. We use t-SNE \cite{van2008visualizing} to visualize feature vectors preceding classifiers, as shown in Figure \ref{fig:3}.

\subsection{Attribute Set Anomaly Detection}

CelebA \cite{liu2015faceattributes} is a large-scale face attributes dataset including 202,599 face images with 40 attribute annotations per image. We randomly sample 9 images with the same 2 attributes and 1 image without those. Each image in CelebA is forced to appear at most once in the sampled dataset. We formulate this problem as a set feature extraction problem because the anomaly is in the minority and most normal members do not deviate much from the set feature. All methods are the same except the set feature extractor part.

\begin{table}[htb]
  \centering
  \begin{tabular}{ccc}
    \toprule
    Method & \#Params(K) & OA(\%)  \\
    \midrule
    Max-pooling & 65.8 & 60.4 $\pm$ 0.44 \\
    Mean-pooling & 65.8 & 65.0 $\pm$ 0.16 \\
    PointNet \shortcite{qi2017pointnet} & 131.6 & 62.3 $\pm$ 0.22 \\
    Deep Sets \shortcite{zaheer2017deep} & 132.6 & 65.0 $\pm$ 0.35 \\
    Set Transformer \shortcite{lee2019set} & 263.4 & 66.4 $\pm$ 0.45 \\
    \midrule
    DuMLP-Pin & 83.3 & \textbf{66.8} $\pm$ 0.43 \\
    \bottomrule
  \end{tabular}
  \caption{The CelebA attribute set anomaly detection results.}
  \label{tab:3}
\end{table}

\begin{figure}[htbp]
  \centering
  \includegraphics[width=0.89\linewidth]{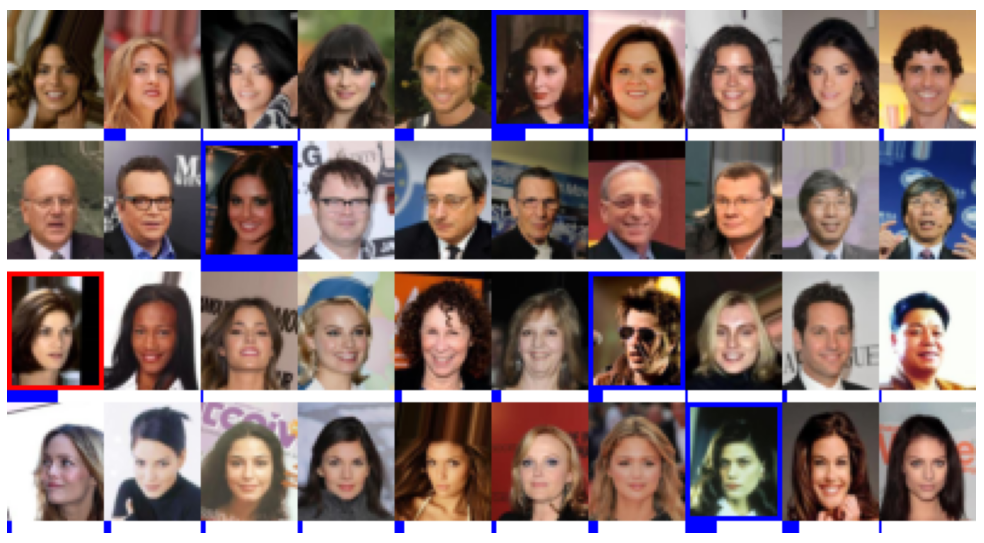}
  \caption{Samples from set anomaly detection datasets, the blue frame denotes anomaly and the red frame denotes wrong-classified image. We visualize the probability by adding a blue bar under each image.}
  \label{fig:6}
\end{figure}

We evaluate all methods under the end-to-end setting without access to the attributes. We use overall accuracy (OA) as the evaluation metric. The structures of other methods are selected from a series of candidates through grid search. From Table \ref{tab:3}, DuMLP-Pin and Set Transformer outperform other methods by a significant margin, but DuMLP-Pin using fewer parameters. Figure \ref{fig:6} shows some samples from the set anomaly detection dataset.

\subsection{Point Cloud Classification}

The ModelNet40 \cite{wu20153d} is a 40-class 3D shape dataset with 9,843 CAD models in the training set and 2,468 CAD models in the test set, respectively. We utilize 1,024 points with normal vectors from the sampled point cloud of \citeauthor{qi2017pointnet++} for fair comparisons. We take overall accuracy as the evaluation metric. All results quoted are taken from the cited papers. 

\begin{table}[htb]
  \centering
  \begin{tabular}{ccc}
    \toprule
    Method & \#Params(M) & OA(\%) \\
    \midrule
    PointNet \shortcite{qi2017pointnet} & 3.48 & 89.2\\
    Deep Sets \shortcite{zaheer2017deep} & 1.34 & 90 $\pm$ 0.3\\
    Set Transformer \shortcite{lee2019set} & 1.14 & 90.4 $\pm$ 1.73\\
    PointNet++ \shortcite{qi2017pointnet++} & 1.74 & 91.9\\
    PCT \shortcite{guo2020pct} & 2.88 & 93.2\\
    Point Transformer \shortcite{zhao2020point} & $\sim$9.6$^*$ & \textbf{93.7}\\
    \midrule
    DuMLP-Pin & \textbf{0.29} & 92.3 $\pm$ 0.12 \\
    \bottomrule
  \end{tabular}
  \caption{The ModelNet40 point cloud classification results. Note that $^*$ indicates that the number of parameters is estimated from unofficial code implementation.}
  \label{tab:2}
\end{table}

\begin{figure}[htb]
  \centering
  \includegraphics[width=0.9\linewidth]{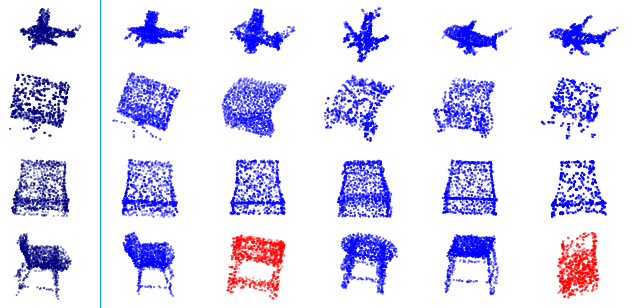}
  \caption{Model retrieval results from the ModelNet40 test set. The left column is the query point cloud, and the others are top-5 retrieved point clouds. Red means incorrect classification.}
  \label{fig:5}
\end{figure}

We also evaluate two models of different sizes, but they have similar performance, meaning that the model is prior to overfitting when adding more parameters. Therefore, we only focus on the small one. The performance of DuMLP-Pin is better than other methods for sets (Deep Sets and Set Transformer) and pioneer methods for point clouds (PointNet and PointNet++), but still 1\%$\sim$2\% worse than state-of-the-art local aggregation methods for point clouds. As a global aggregation method, DuMLP-Pin cannot capture more local information. But the result is competitive since the structure of DuMLP-Pin is simpler and the number of required parameters is remarkably decreased by more than 85\%. We adopt the same model retrieval method as that introduced in \cite{qi2017pointnet} in Figure \ref{fig:5}.

\subsection{Point Cloud Part Segmentation}

The ShapeNetPart \cite{yi2016scalable} is a 3D object part segmentation dataset with 14,006 objects in the training set and 2,874 ones in the test set. It consists of 16 shape categories and 50 part ones, respectively. We utilize 2,048 points like the sampled point clouds of \citeauthor{qi2017pointnet++} for fair comparisons. The mIoU is employed as the evaluation metric and all their results are quoted from the cited papers. 

\begin{table}[htb]
  \centering
  \begin{tabular}{ccc}
    \toprule
    Method & \#Params(M) & mIoU(\%) \\
    \midrule
    PointNet \shortcite{qi2017pointnet} & 7.83 & 83.7\\
    PointNet++ \shortcite{qi2017pointnet++} & 1.30 & 85.1\\
    PCT \shortcite{guo2020pct} & 3.14 & 86.4\\
    Point Transformer \shortcite{zhao2020point} & $\sim$19.2$^*$ & \textbf{86.6}\\
    \midrule
    DuMLP-Pin-S & \textbf{0.70} & 83.5 $\pm$ 0.06 \\
    DuMLP-Pin-L & 0.98 & 84.8 $\pm$ 0.04 \\
    \bottomrule
  \end{tabular}
  \caption{The ShapeNetPart point cloud part segmentation results. Note that $^*$ implies that the data is roughly calculated based on unofficial code implementation.}
  \label{tab:7}
\end{table}

\begin{figure}[htb]
  \centering
  \includegraphics[width=0.9\linewidth]{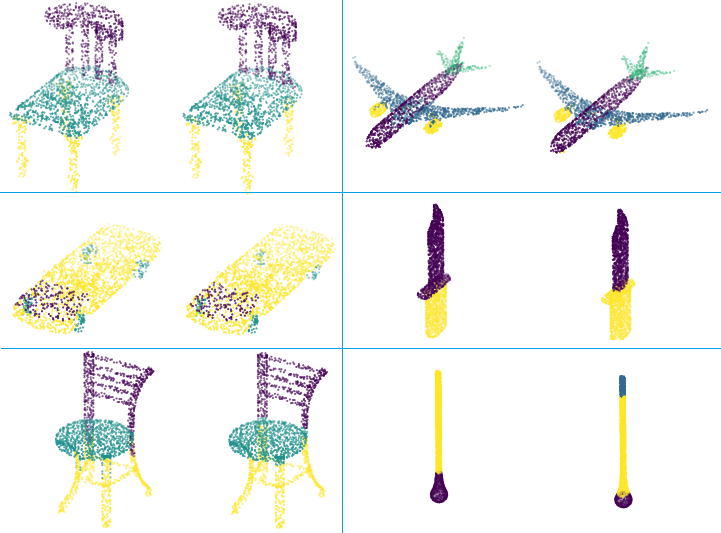}
  \caption{Segmentation results on the ShapeNetPart test set. For each group of point clouds, on the left is the ground truth, while on the right is our prediction.}
  \label{fig:7}
\end{figure}

Like that in previous experiments, we evaluate two models of different sizes: DuMLP-Pin-S with two broadcast blocks and DuMLP-Pin-L with four broadcast blocks. From Table \ref{tab:7}, it is observed that DuMLP-Pin-S somewhat underperforms PointNet and DuMLP-Pin-L is slightly inferior to PointNet++. But DuMLP-Pin has its own advantages. For instance, it does not use any normal vectors as input. Without the use of local features, it is reasonable that the performance of DuMLP-Pin is only 2\% lower than that of attention-based local aggregation methods for point cloud problems. We visualize some of the segmentation examples obtained by DuMLP-Pin in Figure \ref{fig:7}.

\subsection{Computational Complexity Analysis}

We compare the computational complexity of different methods in the point cloud classification task, as listed in Table \ref{tab:9}. It is readily observed that DuMLP-Pin has the lowest number of parameters and MACs per sample among all the methods. There are two reasons behind such minimum time-consuming advantage. First, the proposed method has smaller hidden layer dimensions. Second, it also has linear complexity with respect to the number of points, while the computational complexity of transformer-based methods is generally quadratic.
\begin{table}[htbp]
	\centering
	\begin{tabular}{ccc}
		\toprule
		Method & \#Params(M) & MACs/sample(M) \\
		\midrule
		PointNet & 3.48 & 440 \\
		Deep Sets & 1.34 & 135 \\
		Set Transformer & 1.14 & 1,284 \\
		PointNet++ & 1.74 & 3,932 \\
		PCT & 2.88 & 2,286 \\
		Point Transformer & $\sim$9.6$^*$ & $\sim$18,314$^*$ \\
		\midrule
		DuMLP-Pin & \textbf{0.29} & \textbf{19} \\
		\bottomrule
	\end{tabular}
	\caption{The comparison of time efficiency among all the methods in the point cloud classification. $^*$ indicates that the value is estimated from unofficial code implementation.}
	\label{tab:9}
\end{table}

\subsection{Ablation Study}
\label{sec:1}

The analysis of the influence of several factors including nonlinear activation function, factorization, and decomposition order are further conducted. All experiments are done with ModelNet40, where all MLPs adopted in DuMLP-Pin contain three hidden layers with sizes of 32, 128, and $k$ which may vary in different experiments, respectively.

\begin{figure}[htbp]
  \centering
  \includegraphics[width=0.8\linewidth]{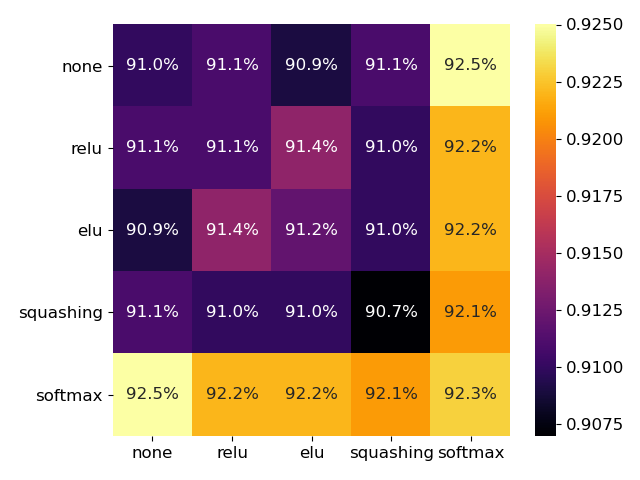}
  \caption{Influence of nonlinear activation function.}
  \label{fig:4}
\end{figure}

\textbf{Nonlinear Activation Function.} All models are the same except the activation functions of two MLPs. For set-wise activation functions, we adopt squashing introduced in \cite{sabour2017dynamic} besides softmax. Figure \ref{fig:4} visualizes different activation functions used in our work. It can be seen that softmax dominates the table with all metrics around 1\% greater than their non-softmax counterparts. Squashing seems to lose so much information that it has poor performance. Generally, we prefer using two softmax or one softmax and the other no activation.

\textbf{Factorization.} By controlling the dimension of feature vectors as 1,024, we experiment on several different factorizations. The influence of factorization does not seem critical because the OA of the best model is only 0.7\% larger than the worst model in Table \ref{tab:6}. However, since the square factorization ($32\times 32$) needs the fewest parameters, it is highly recommended to use square factorization first and try other factorizations as well if possible. To control the variable, we use both activation functions as softmax, which makes the result slightly different from that with one single softmax in Figure \ref{fig:4}.

\begin{table}[htbp]
  \centering
  \begin{tabular}{ccc}
    \toprule
    Factorization & \#Params(K) & OA(\%) \\
    \midrule
    $1\times 1,024$ & 143.8 & 91.8 \\
    $2\times 512$ & 76.9 & 92.1 \\
    $4\times 256$ & 43.6 & 91.6 \\
    $8\times 128$ & 27.4 & 91.7 \\
    $16\times 64$ & 20.0 & 91.9 \\
    $32\times 32$ & \textbf{17.9} & \textbf{92.3} \\
    \bottomrule
  \end{tabular}
  \caption{Influence of factorization.}
  \label{tab:6}
\end{table}

\textbf{Decomposition Order.} By setting $N=1,024$ and controlling the dimension of feature vectors as 1,024, the decomposition always exists according to Theorem \ref{the:2}. Because multiple softmax makes the value too small, we only use one softmax in all experiments except the first one (since it all degenerates to one). From Table \ref{tab:5}, the performance of order 2 is better than all the other methods. The possible reason is that the use of high-order decomposition is equivalent to adding stronger constraints. In this case, it is hard to use any vanilla MLPs to parameterize each function in the decomposition.

\begin{table}[htbp]
  \centering
  \begin{tabular}{cccc}
    \toprule
    Order & Factorization & \#Params(K) & OA(\%) \\
    \midrule
    1 & $1,024$ & 138.9 & 90.3 \\
    2 & $32^2$ & \textbf{17.9} & \textbf{92.5} \\
    3 & $16\times 8^2$ & 18.5 & 92.0 \\
    4 & $8^2\times 4^2$ & 22.2 & 91.6 \\
    5 & $4^5$ & 26.5 & 91.2 \\
    10 & $2^{10}$ & 50.3 & 91.3 \\
    \bottomrule
  \end{tabular}
  \caption{Influence of decomposition order.}
  \label{tab:5}
\end{table}

\section{Conclusion}

In this paper, we propose a novel global aggregation permutation-invariant network based on dual MLP dot-product decomposition, which is called DuMLP-Pin. We strictly prove that two or more permutation-equivariant functions can be leveraged to set up one permutation-invariant function. The high efficiency of DuMLP-Pin is also analyzed by converting the optimization problem of DuMLP-Pin to that of Deep Sets with strong constraints. Finally, we evaluate the performance of our DuMLP-Pin on different tasks. The experimental results illustrate that DuMLP-Pin achieves the best results on the pixel set classification and the attribute set anomaly detection. Additionally, the performance of DuMLP-Pin is only slightly lower than existing SOTA methods, while the number of required parameters is significantly reduced. In the future, we hope the DuMLP-Pin could extensively be exploited as a competitive global aggregation baseline for a wide range of unordered or unstructured data problems in terms of its distinguishing performance.

\section*{Acknowledgments}

This work was supported by Alibaba Group through Alibaba Innovative Research Program. It was also supported in part by the National Science Foundation of China (NSFC) under Grant No. 62176134 and by a grant from the Institute Guo Qiang (2019GQG0002), Tsinghua University.

\bibliography{aaai22}

\end{document}


\maketitle

\appendix

\section{Theoretical Analysis}
\label{apx:1}

\begin{lemma}[MDD]
  \label{lem:2}
  $\forall A\in\mathbb{R}^{m\times n}, m\leq n, \forall l\geq m, \forall B$ $\in\{X\in\mathbb{R}^{m\times l}\ \big|\ \mbox{rank}(X)=m\}$. Let $S=\left\{C\in\mathbb{R}^{l\times n}|\right.$ $\left.BC=A\right\}$, then
  \begin{equation}
    S\neq \Phi\text{ and }S=\left\{C_p+X_h\Lambda|\Lambda\in\mathbb{R}^{(l-m)\times n}\right\}
    \label{equ:7},
  \end{equation}
  where $C_p$ stands for a particular solution of $BC=A$, and the columns of $X_h$ are basis of Ker$(B)$.
  \label{lem:1}
\end{lemma}

\textbf{Proof.} We will prove this lemma by solving $C$ from $BC=A$.

$\forall B\in\{X\in\mathbb{R}^{m\times l}|\mbox{rank}(X)=m\}$, let us consider the following equation:
\begin{equation}
  BX=Y,
  \label{equ:1}
\end{equation}
where $X\in\mathbb{R}^l$ and $Y\in\mathbb{R}^m$. From linear algebra, the solution of Equation \ref{equ:1} always exists, and can be written as the sum of a particular solution $C_p$ and a homogenous solution $X_h\in\text{Ker}(B)$. We can also derive that $\text{dim}\left[\text{Ker}\left(B\right)\right]=l-m$.

Partition $A$ and $C$ into a sequence of column vectors: $A=\left[A_1,\cdots,A_n\right]$ and $C=\left[C_1,\cdots,C_n\right]$, respectively, we have
\begin{equation}
  BC_i=A_i,i=1,\cdots,n\Rightarrow C_i=X_{p,i}+\sum_{j=1}^{l-m}\lambda_{ji}X_{h,j},
\end{equation}
where $X_{p,i}$ is a particular solution and $X_{h,j}$ is the $j$-th element of the basis of $\text{Ker}(B)$. Finally, $C$ can be constructed by concatenating all $C_i$, i.e.
\begin{equation}
  C=C_p+X_h\Lambda,
\end{equation}
where $C_p=\left[X_{p,1},\cdots,X_{p,n}\right]\in\mathbb{R}^{l\times n},X_h=\left[X_{h,1},\cdots,X_{h,l-m}\right]\in\mathbb{R}^{l\times(l-m)}$, and $\Lambda=(\lambda_{ji})\in\mathbb{R}^{(l-m)\times n}$. Note that $\Lambda$ controls the degrees of freedom of $S=\left\{C_p+X_h\Lambda|\Lambda\in\mathbb{R}^{(l-m)\times n}\right\}$. 

\begin{theorem}
  \label{the:3}
  For any permutation-invariant function $f:\mathbb{R}^{N\times p}\rightarrow\mathbb{R}^{s\times t}$, there exist two permutation-equivariant functions $g^{(1)}:\mathbb{R}^{N\times p}\rightarrow\mathbb{R}^{N\times s}, g^{(2)}:\mathbb{R}^{N\times p}\rightarrow\mathbb{R}^{N\times t}$, such that
  \begin{equation}
    \forall X\in\mathbb{R}^{N\times p},f(X)=\left(g^{(1)}(X)\right)^Tg^{(2)}(X)
  \end{equation}
  \begin{equation*}
    \text{if and only if } N\geq\min\{s,t\}.
  \end{equation*}
\end{theorem}

\textbf{Proof.} Without loss of generality, let us assume $s\leq t$. 

\textbf{\emph{Sufficiency.}} The core is to prove that there exists a permutation-equivariant function satisfying $\text{rank}\left[g^{(1)}(X)\right]\equiv s$.

Suppose that $E=\mathbb{R}^{N\times p}$ and $F=\left\{X\in\mathbb{R}^{N\times s}|\text{rank}\left(X\right)=s\right\}$. Let us introduce an equivalence relation $R: aRb\Leftrightarrow \exists P\in\mathcal{P}_N$ such that $a=Pb$ and $E/R$ indicates the quotient set. First, we define the surjective map $\psi:E\rightarrow E/R$, which maps each element in $E$ to its equivalence class with respect to $R$. Second, the injective map $\varphi:E/R\rightarrow E$, which maps each equivalence class to its representative, is defined. Finally, let us introduce the representative set $G=\left\{X\in E|\exists Y\in E/R,\varphi(Y)=X\right\}$. For all mappings $\phi:G\rightarrow F$, we can construct a permutation-equivariant function $g_\phi:E\rightarrow F$ based on $\phi$:
\begin{equation}
  g_\phi(X)=\left\{\begin{aligned}
    \phi(X) &\ \ \ \ X\in G,\\
    P^T\phi\left(PX\right) & \ \ \ \ X\notin G, PX\in G.
  \end{aligned}\right.
  \label{equ:3}
\end{equation}

Let us denote $g^{(1)}=g_\phi$. According to Lemma \ref{lem:2}, for each $g^{(1)}(X)$, we can derive corresponding $g^{(2)}(X)$ so that $f(X)=\left(g^{(1)}(X)\right)^Tg^{(2)}(X)$. Considering that $f(X)$ is permutation-invariant and $g^{(1)}(X)$ is permutation-equivariant, $g^{(2)}(X)$ is also permutation-equivariant.

\textbf{\emph{Necessity.}} Due to the arbitrariness of $f$, we choose $X$ such that $\text{rank}\left[f(X)\right]=s$. It follows:
\begin{equation}
  \begin{aligned}
    s&=\text{rank}\left[f(X)\right]=\text{rank}\left[\left(g^{(1)}(X)\right)^Tg^{(2)}(X)\right]\\
    &\leq\text{rank}\left[g^{(1)}(X)\right]\leq N.
  \end{aligned}
\end{equation}

\begin{theorem}
  \label{the:2}
  For any permutation-invariant function $f:\mathbb{R}^{N\times p}\rightarrow\mathbb{R}^{c_1\times\cdots\times c_n}$, there exist $n$ permutation-equivariant functions $g^{(1)}:\mathbb{R}^{N\times p}\rightarrow\mathbb{R}^{N\times c_1}, \cdots, g^{(n)}:\mathbb{R}^{N\times p}\rightarrow\mathbb{R}^{N\times c_n}$, such that ($1\leq a_j\leq c_j$)
  \begin{equation}
    \forall X\in\mathbb{R}^{N\times p},f(X)_{a_1,\cdots,a_n}=\sum_{i=1}^N\prod_{j=1}^n\left(g^{(j)}(X)\right)_{i,a_j}
    \label{equ:2}
  \end{equation}
  \begin{equation*}
    \text{if }N\geq\frac{\prod_{j=1}^nc_j}{\max_jc_j}\text{ and only if }N\geq\frac{\prod_{j=1}^nc_j}{\sum_{j=1}^nc_j}
  \end{equation*}
  \begin{equation*}
    \text{and }N\geq\max_{C\subset\left\{c_1, \cdots, c_n\right\}}\min\left\{\prod_{c\in C}c, \prod_{c\in\overline{C}}c \right\}.
  \end{equation*}
\end{theorem}

\textbf{Proof.} Without loss of generality, let us assume $c_1\leq\cdots\leq c_n$. 

\textbf{\emph{Sufficiency.}} First of all, make flattening of all dimensions of $f(X)$ except for the last one. The decomposition in Equation \ref{equ:2} could be rewritten in a matrix form. Here we take $\prod_{k=n}^{n-1}c_k=1$.
\begin{equation}
  f^\prime(X)=\left[g^\prime(X)\right]^Tg^{(n)}(X),
\end{equation}
\begin{align*}
  &\text{where } f^\prime(X)\in\mathbb{R}^{\prod_{j=1}^{n-1}c_j\times c_n}, g^\prime(X)\in\mathbb{R}^{N\times \prod_{j=1}^{n-1}c_j}\\
  &f^\prime(X)_{\sum_{j=1}^{n-1}(a_j-1)\prod_{k=j+1}^{n-1}c_k+1,a_n}=f(X)_{a_1,\cdots,a_n}\\
  &g^\prime(X)_{i,\sum_{j=1}^{n-1}(a_j-1)\prod_{k=j+1}^{n-1}c_k+1}=\prod_{j=1}^{n-1}\left(g^{(j)}(X)\right)_{i,a_j}.
\end{align*}

Suppose that $F^\prime=\left\{X\in\mathbb{R}^{N\times \prod_{j=1}^{n-1}c_j}|\text{rank}\left(X\right)=\right.$ $\left.\prod_{j=1}^{n-1}c_j\text{ and }\exists V^{(j)}\in\mathbb{R}^{N\times c_j},x_{i,\sum_{j=1}^{n-1}(a_j-1)\prod_{k=j+1}^{n-1}c_k+1}\right.$ $\left.=\prod_{j=1}^{n-1}v^{(j)}_{i,a_j}\right\}\neq\Phi$. Following the same procedure in the proof of Theorem \ref{the:3}, the decomposition can be done. Next, we constuct an element $X$ in $F^\prime$ to prove that $F^\prime$ is not an empty set.

Choose $v^{(j)}_{i,a_j}=\exp\left((i-1)(a_j-1)\prod_{k=j+1}^{n-1}c_k\right)$, then it has
\begin{equation}
  \begin{aligned}
    &x_{i,\sum_{j=1}^{n-1}(a_j-1)\prod_{k=j+1}^{n-1}c_k+1}=\prod_{j=1}^{n-1}v^{(j)}_{i,a_j}\\
    =&\prod_{j=1}^{n-1}\exp\left((i-1)(a_j-1)\prod_{k=j+1}^{n-1}c_k\right)\\
    =&\exp\left((i-1)\sum_{j=1}^{n-1}(a_j-1)\prod_{k=j+1}^{n-1}c_k\right).
  \end{aligned}
\end{equation}

By introducing $d=\sum_{j=1}^{n-1}(a_j-1)\prod_{k=j+1}^{n-1}c_k+1$, we have:
\begin{equation}
    x_{i,d}=\exp\left[(i-1)(d-1)\right]=\left[\exp\left(i-1\right)\right]^{d-1}.
\end{equation}

Therefore, $X$ is a \textbf{Vandermonde matrix} with $N$ distinct basis elements of $e^0,e^1,\cdots,e^{N-1}$. Considering $N\geq\prod_{j=1}^{n-1}c_j$, according to the property of Vandermonde matrix, we have that $\text{rank}\left(X\right)=\prod_{j=1}^{n-1}c_j$. Thus $F^\prime$ is not an empty set.

Assume that $E=\mathbb{R}^{N\times p}$. Let us introduce an equivalence relation $R: aRb\Leftrightarrow \exists P\in\mathcal{P}_N, a=Pb$ and the quotient set $E/R$. We continue to define the surjective map $\psi:E\rightarrow E/R$, which maps each element in $E$ to its equivalence class with respect to $R$, and the injective map $\varphi:E/R\rightarrow E$, which maps each equivalence class to its representative. Meanwhile, suppose that the representative set $G=\left\{X\in E|\exists Y\in E/R,\varphi(Y)=X\right\}$.

Since $F^\prime$ is not an empty set, $\phi^\prime(X)$ can be decomposed into a series of functions: $\phi^{(1)}:G\rightarrow\mathbb{R}^{N\times c_1}, \cdots, \phi^{(n-1)}:G\rightarrow\mathbb{R}^{N\times c_{n-1}}$ for all mappings $\phi^\prime:G\rightarrow F^\prime$. By use of Equation \ref{equ:3}, each $\phi^{(j)}$ and $\phi^\prime$ can be utilized to build $g^{(j)}$ and $g^\prime$, respectively. Eventually, according to Lemma \ref{lem:2}, we can make derivation of corresponding $g^{(n)}(X)$ for each $g^\prime(X)$, which is permutation-equivariant.  

\textbf{\emph{Necessity.}} Equation \ref{equ:2} may be regarded as multivariate polynomial equations with $\prod_{j=1}^nc_j$ independent equations and $N\sum_{j=1}^nc_j$ variables. To make sure the solutions exist, the number of equations should not be greater than the number of variables:
\begin{equation}
  \prod_{j=1}^nc_j\leq N\sum_{j=1}^nc_j\Rightarrow N\geq\frac{\prod_{j=1}^nc_j}{\sum_{j=1}^nc_j}.
\end{equation}

We may also transfer an $n$D tensor into a matrix by reordering and grouping some dimensions. Assume that we group $C\subset\left\{c_1,\cdots,c_n\right\}$ dimensions and get corresponding matrix $f^C(X)$. Due to the arbitrariness of $f$, we choose $X$ such that $f^C(X)$ is full-rank, it has:
\begin{equation}
  \begin{aligned}
    &\min\left\{\prod_{c\in C}c, \prod_{c\in\overline{C}}c\right\}=\text{rank}\left[f^C(X)\right]\\
    =\ &\text{rank}\left[\left(g^{C}(X)\right)^Tg^{\overline{C}}(X)\right]\leq\text{rank}\left[g^{C}(X)\right]\leq N.
  \end{aligned}
\end{equation}

Owing to the arbitrariness of $C$, it is easy to make derivation below:
\begin{equation}
  N\geq\max_{C\subset\left\{c_1, \cdots, c_n\right\}}\min\left\{\prod_{c\in C}c, \prod_{c\in\overline{C}}c \right\}.
\end{equation}

Specifically, we construct $g^{(1)}(X)$ through MLPs. It is almost impossible to ensure that it always outputs full-rank matrices for all inputs. We tackle this problem mainly from two perspectives: First, we seldom observe rank-deficient outputs, especially when $s\ll N$, in experiments. This is because rank-deficient states are more \textbf{unstable} than full-rank ones. Second, let us have constraints that $\max\left\{s,t\right\}\leq N$. Thus we could treat $g^{(2)}(X)$ as $g^{(1)}(X)$ if $\text{rank}\left[g^{(1)}(X)\right]<s$. In other words, there occur such problems as both $\text{rank}\left[g^{(1)}(X)\right]<s$ and $\text{rank}\left[g^{(2)}(X)\right]<t$ hold. Actually, it is a rare case. We formulate the \textbf{unstable} property as Lemma \ref{lem:3}.
\begin{lemma}
  \label{lem:3}
  For $\forall Y\in\mathbb{R}^{N\times s}(N\geq s),\ \text{rank}(Y)<s$ and for $\forall\varepsilon>0,\ \exists\Delta_0\in\mathbb{R}^{N\times s}$ that makes $||\Delta_0||_F<\varepsilon$, such that
  \begin{equation}
    \text{rank}(Y+\Delta_0)=s.
    \label{equ:8}
  \end{equation}
  For $\forall Y\in\mathbb{R}^{N\times s}(N\geq s),\ \text{rank}(Y)=s$ and $\exists\varepsilon_0>0$ that makes $\forall\Delta\in\mathbb{R}^{N\times s},\ ||\Delta||_F<\varepsilon$, such that
  \begin{equation}
    \text{rank}(Y+\Delta)=s.
    \label{equ:9}
  \end{equation}
  where $||\cdot||_F$ denotes the Frobenius norm.
\end{lemma}

\textbf{Proof.} Let the first $s$ rows of $Y$ be indicated by $Y_s$. For Equation \ref{equ:8}, we introduce $f(x)$ that satisfies:
\begin{equation}
  f(x)=\det\left(Y_s+xI\right),
\end{equation}
where $f(x)$ is an $s$-order polynomial with respect to $x$, having at most $s$ different roots in $\mathbb{R}$, and $\text{rank}(Y_s)\leq\text{rank}(Y)<s\Rightarrow f(0)=0$. Except for these roots, $f(x)\neq 0$. Therefore, $\exists\epsilon>0$ such that $f(x)\neq 0$ for $\forall x\in(0,\epsilon)$. 

In summary, if we take $x_0=\dfrac{1}{2}\min\left\{\epsilon,\ \dfrac{\varepsilon}{\sqrt{s}}\right\}\ \Delta_0=x_0\begin{bmatrix}I\\0\end{bmatrix}$ for $\forall\varepsilon>0$, it follows:
$$x_0\in(0,\epsilon)\Rightarrow f(x_0)\neq 0\Rightarrow\text{rank}(Y_s+x_0I)=s$$
$$\Rightarrow\text{rank}\left(Y+\Delta_0\right)=s,$$
\begin{equation}
  ||\Delta_0||_F=x_0\Bigg|\Bigg|\begin{bmatrix}I\\0\end{bmatrix}\Bigg|\Bigg|_F=\sqrt{s}x_0<\varepsilon.
\end{equation}

For Equation \ref{equ:9}, we introduce $h(x)$ described as follows:
\begin{equation}
  h(\Delta_s)=\det\left(Y_s+\Delta_s\right), \Delta_s\in\mathbb{R}^{s\times s},
\end{equation}
where $\Delta_s$ denotes the first $s$ rows of $\Delta$ and $h(\Delta_s)$ is a continuous function with respect to each element in $\Delta_s$. $h(\bm{0})\neq 0\Rightarrow \exists\epsilon>0$, such that $||\Delta_s||_F<\epsilon\Rightarrow h\left(\Delta_s\right)\neq 0$.

In conclusion, if we take $\varepsilon_0=\epsilon$, it has:
\begin{equation}
  \begin{aligned}
    &||\Delta||_F<\varepsilon_0\Rightarrow||\Delta_s||_F<\varepsilon_0\Rightarrow h\left(\Delta_s\right)\neq 0\\
  \Rightarrow&\ \text{rank}\left(Y_s+\Delta_s\right)=s\Rightarrow\text{rank}\left(Y+\Delta\right)=s.
  \end{aligned}
\end{equation}

\section{Experimental Setting}
\label{apx:2}

All layers in MLPs except the last layer in classification tasks have structures with the template linear + batch normalization + activation function, where all activation functions are ReLU. Dropout is only placed prior to the output of the aggregation block with dropout ratio of 0.1. We use MLP$[d_1,\cdots,d_n]$ to represent MLPs with the input dimension $d_1$, the output dimension $d_n$, and the hidden layer dimensions of $d_2,\cdots,d_{n-1}$. In our experiments, the stochastic gradient descent (SGD) with 0.9 momentum and 0.0001 weight decay is adopted as the optimizer. All experiments are done on one single RTX 3090.

\subsection{Pixel Set Classification}

\begin{figure}[htbp]
  \centering
  \includegraphics[width=0.75\linewidth]{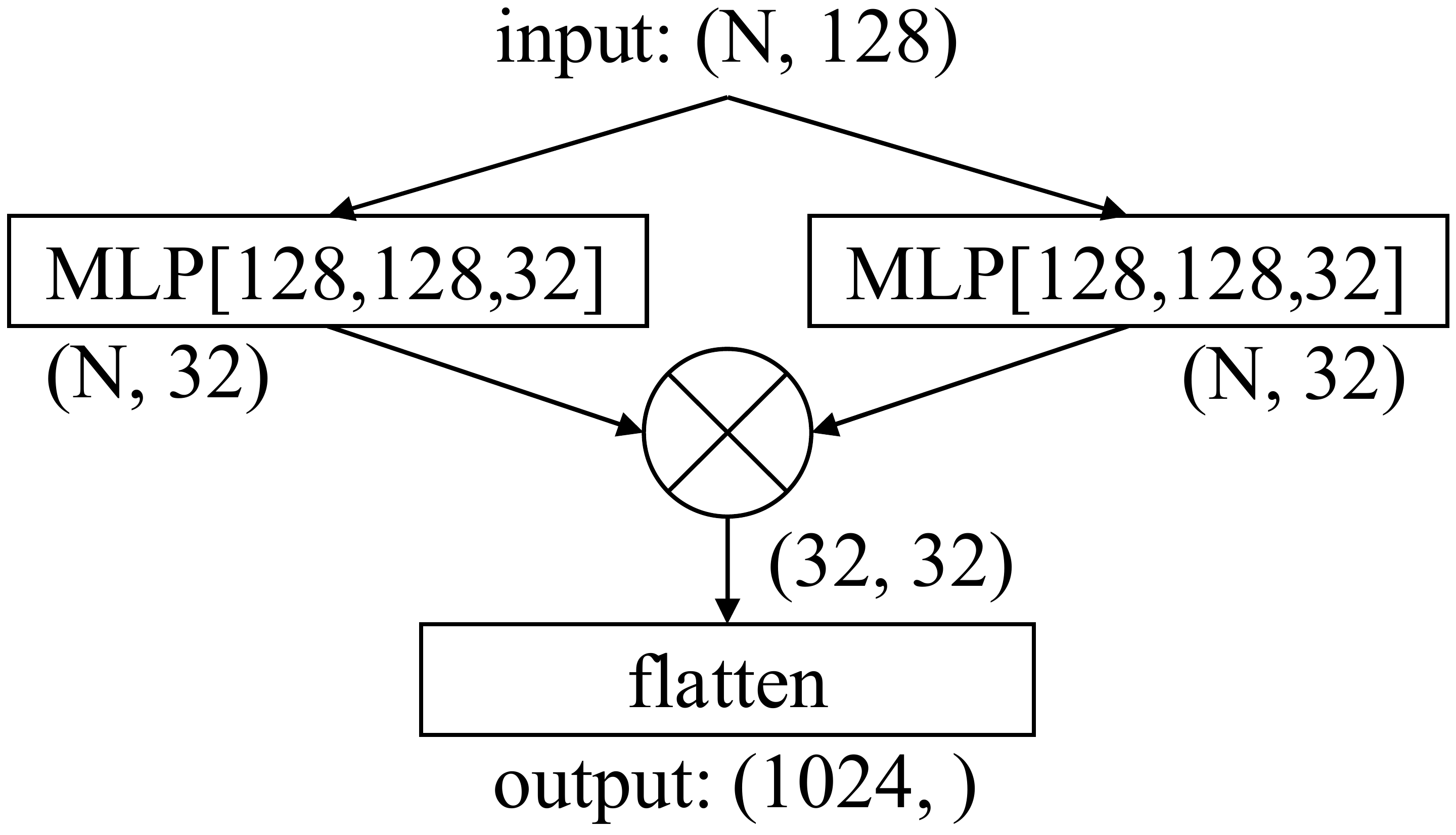}
  \centerline{(b) Aggregation block II}
  \includegraphics[width=0.75\linewidth]{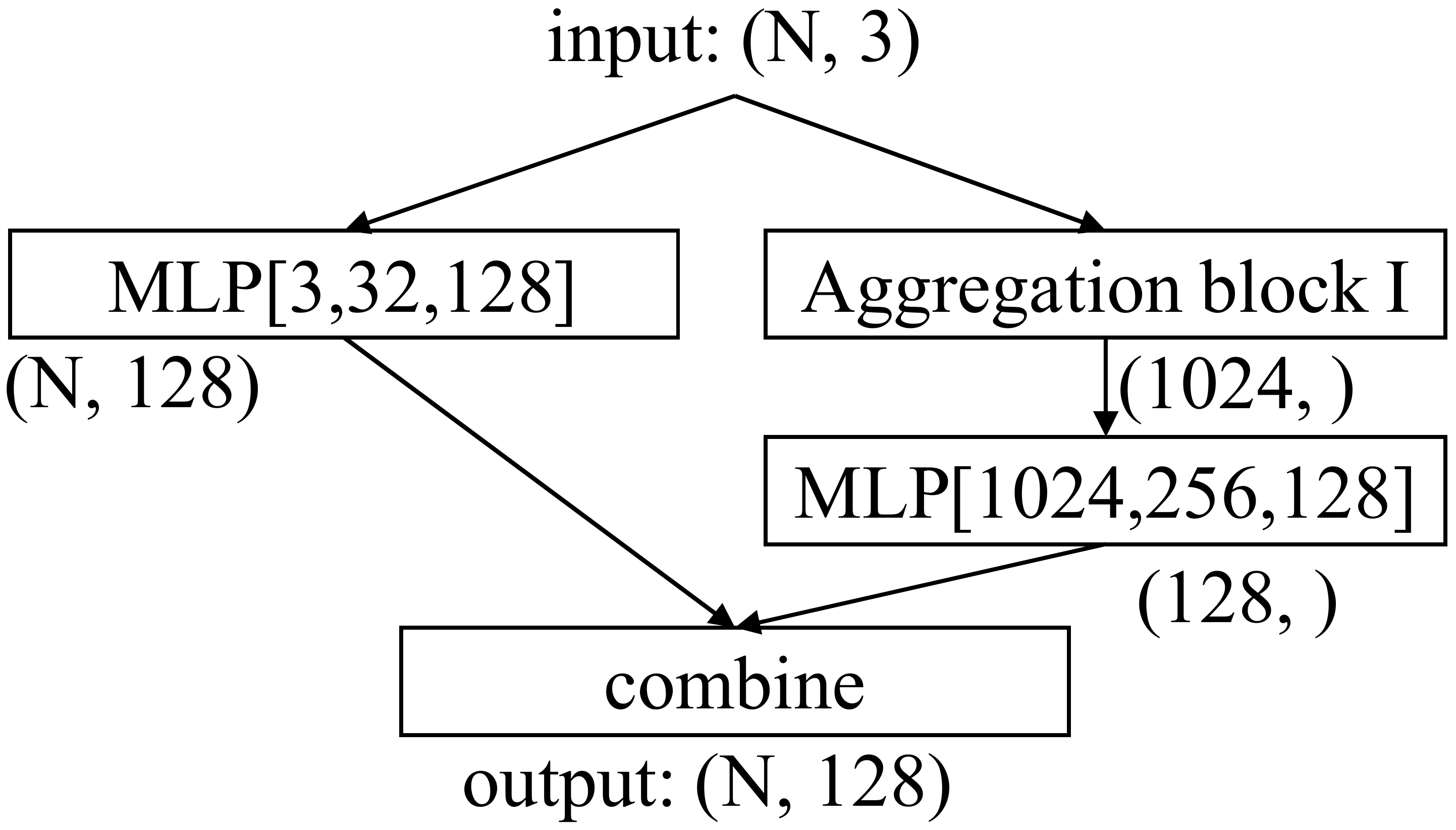}
  \centerline{(c) Broadcast block I}
  \includegraphics[width=0.75\linewidth]{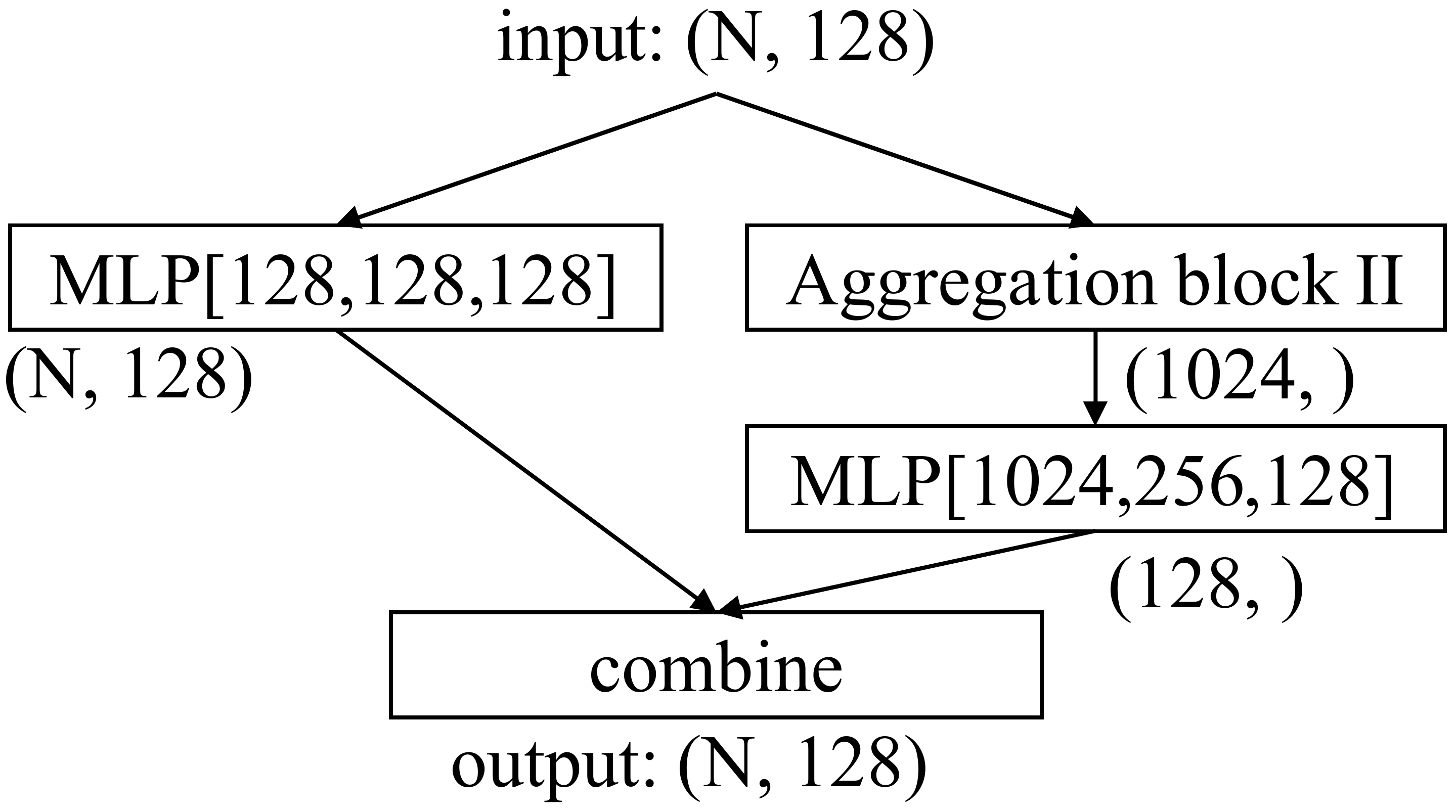}
  \centerline{(d) Broadcast block II}
  \includegraphics[width=0.37\linewidth]{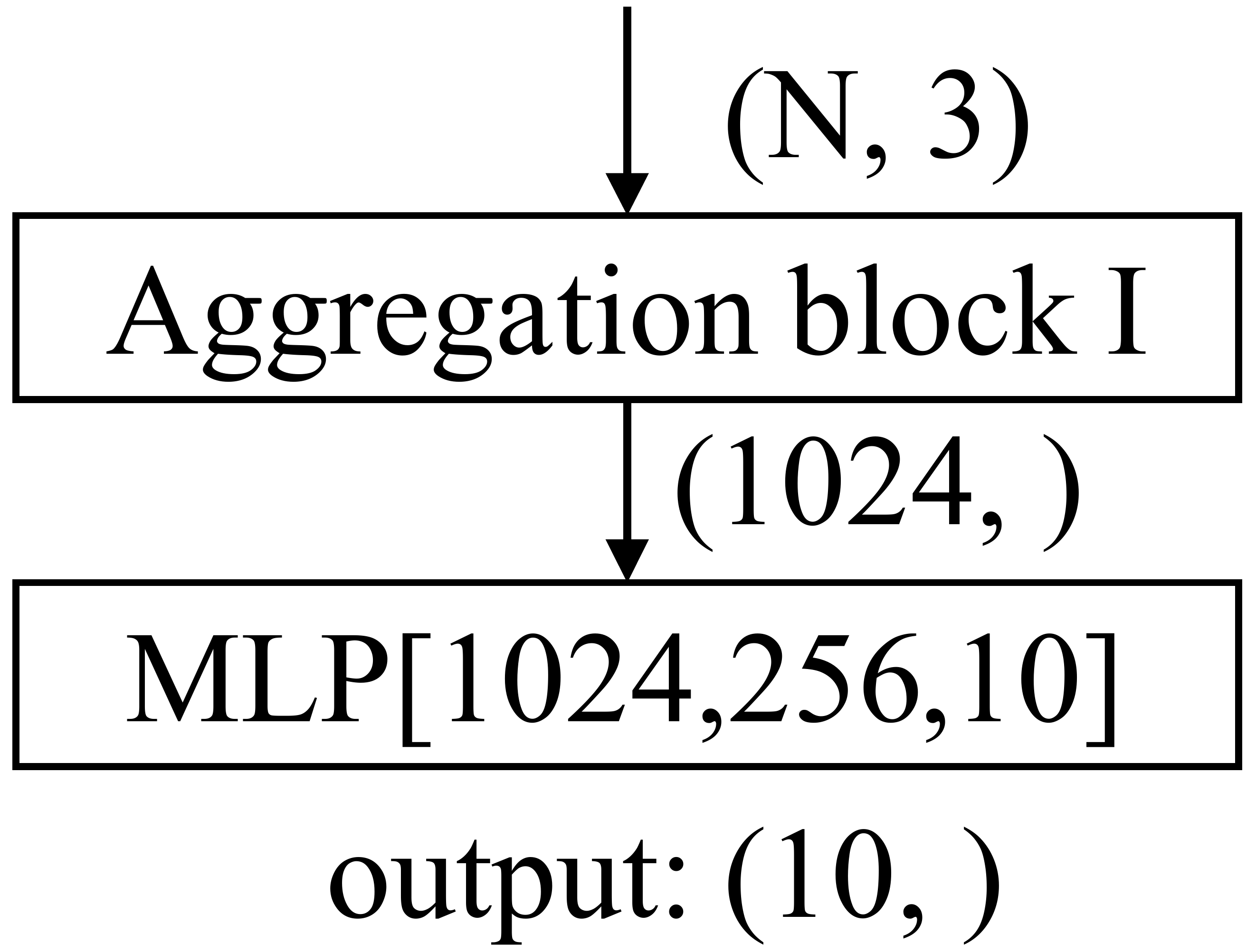}
  \centerline{(e) DuMLP-Pin-S}
  \includegraphics[width=0.4\linewidth]{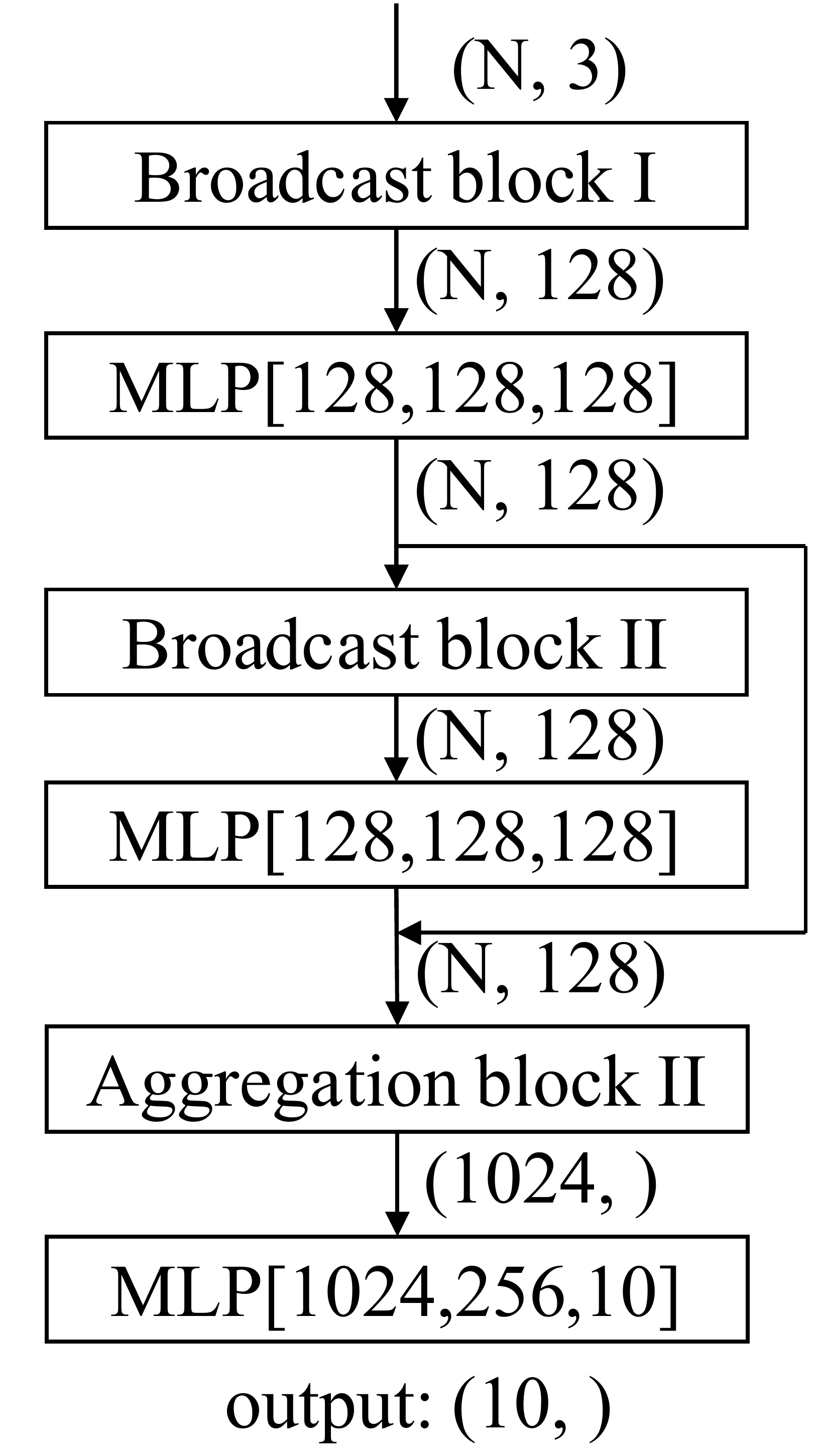}
  \centerline{(f) DuMLP-Pin-L}
  \caption{Detailed structure diagrams of DuMLP-Pin-S and DuMLP-Pin-L used in MNIST pixel set classification.}
  \label{fig:6}
\end{figure}

Let $M\in\mathbb{R}^{28\times 28}$ and $S\in\mathbb{R}^{784\times 3}$ represent an image and its corresponding set, respectively. Each row of $S$ has three elements, where the first two ones denote the relative coordinates in $\left[-1, 1\right]$ and the third one indicates the pixel gray value. The structure diagrams of both DuMLP-Pin-S and DuMLP-Pin-L are shown in Figure \ref{fig:6}.

\begin{figure}[htbp]
  \centering
  \includegraphics[width=0.75\linewidth]{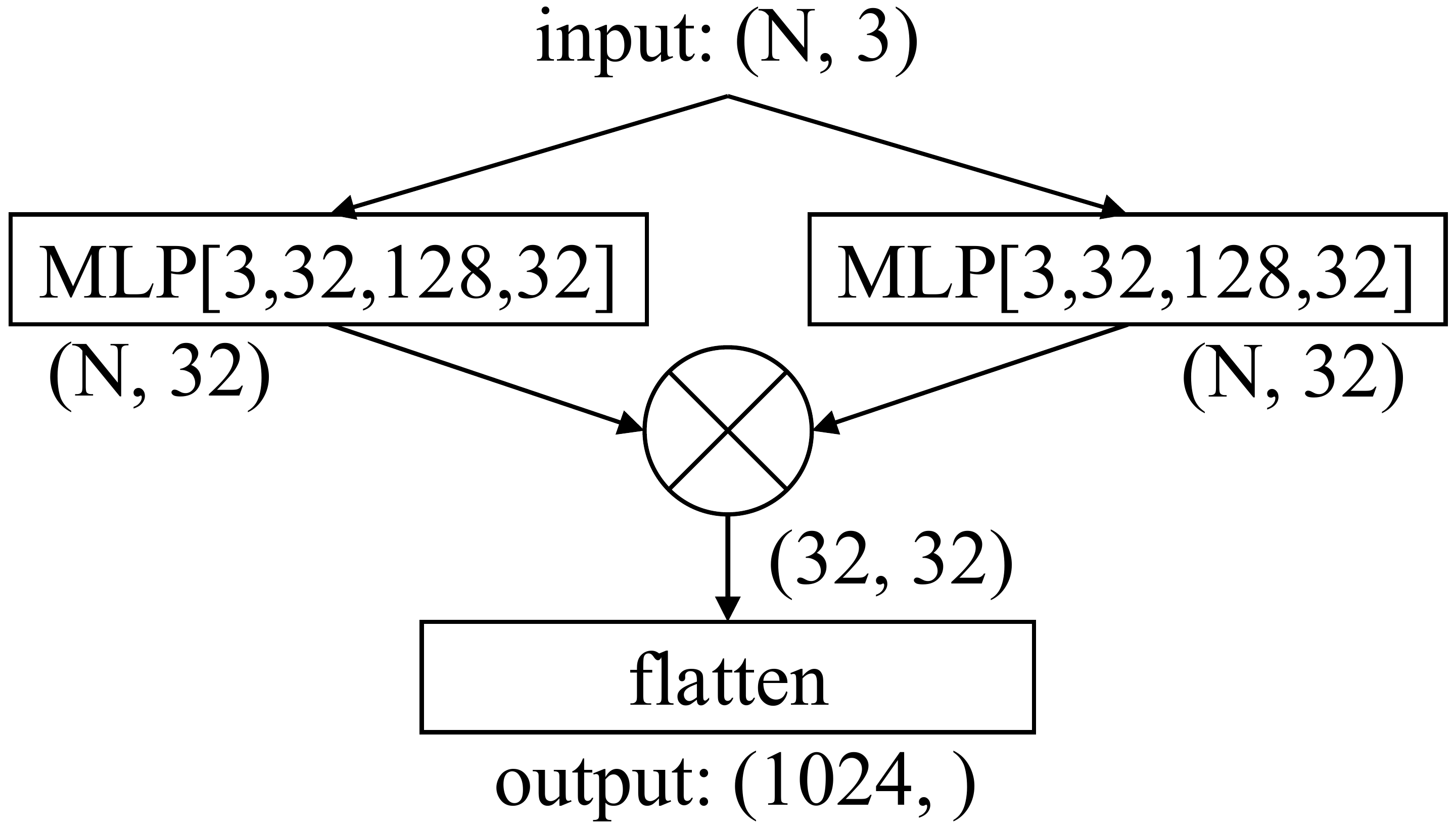}
  \centerline{(a) Aggregation block I}
\end{figure}

Note that (a),(b),(c), and (d) are referred to as 'classes' instead of 'instances', meaning that no parameters are shared in DuMLP-Pin-L. The batch size is set to 32. We train the two models for 250 epochs. The initial learning rate is set to 0.01 and is dropped by 10x at epoch 200. We utilize two softmax as activation functions in both MLPs.

\subsection{Attribute Set Anomaly Detection}

\begin{figure}[h]
  \centering
  \includegraphics[width=0.43\linewidth]{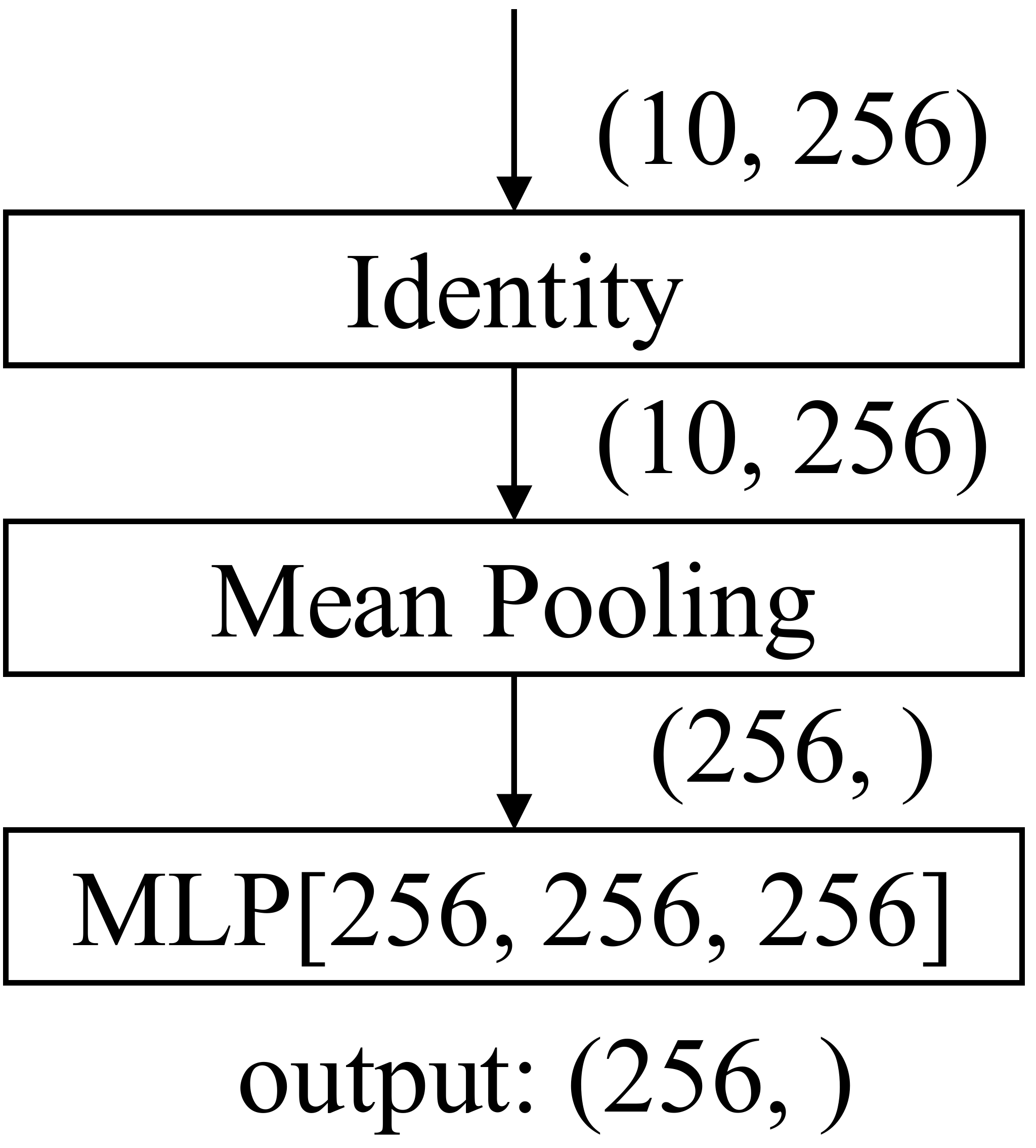}
  \centerline{(d) Deep Sets}
  \includegraphics[width=0.43\linewidth]{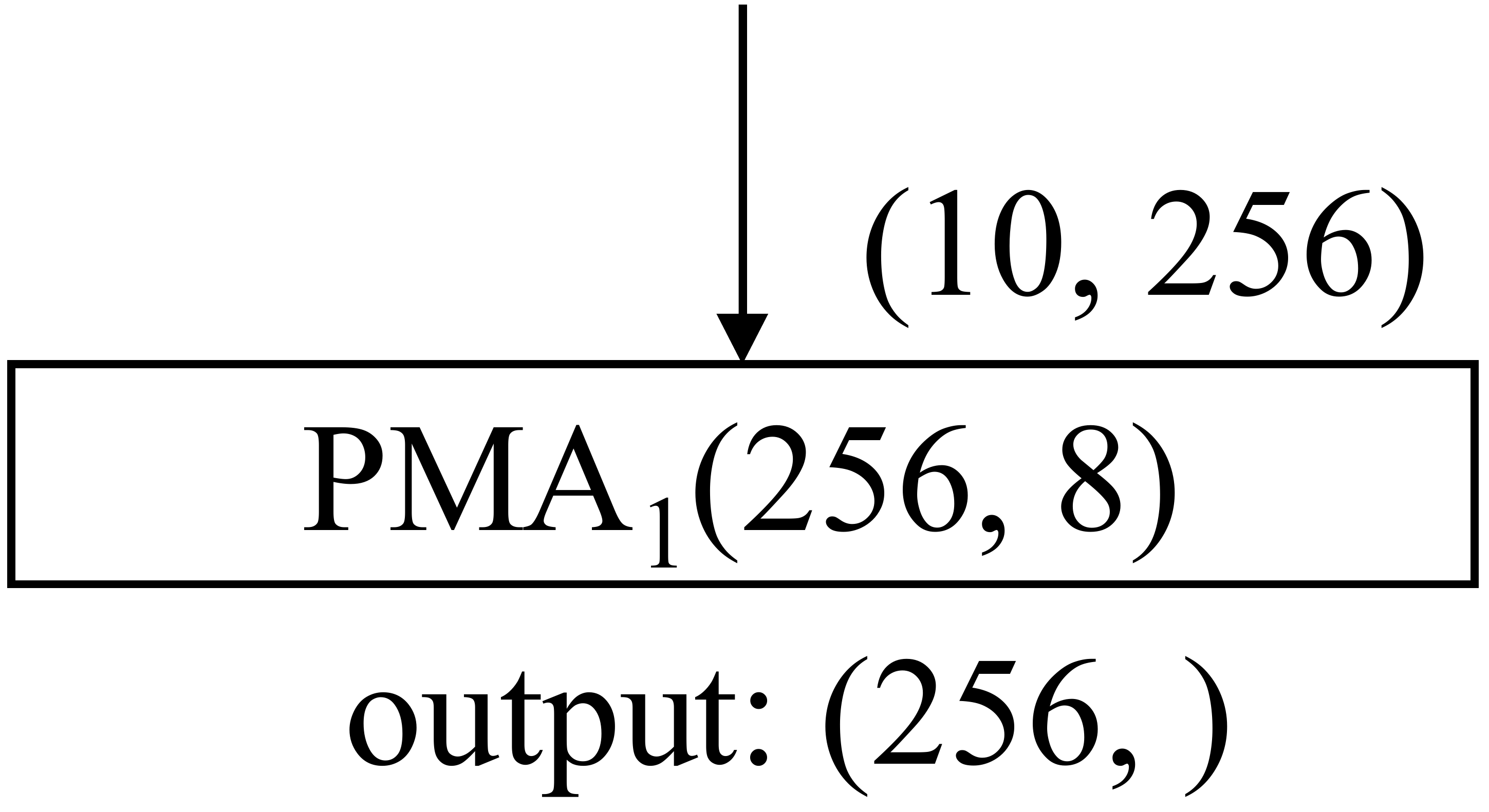}
  \centerline{(e) Set Transformer}
  \includegraphics[width=0.75\linewidth]{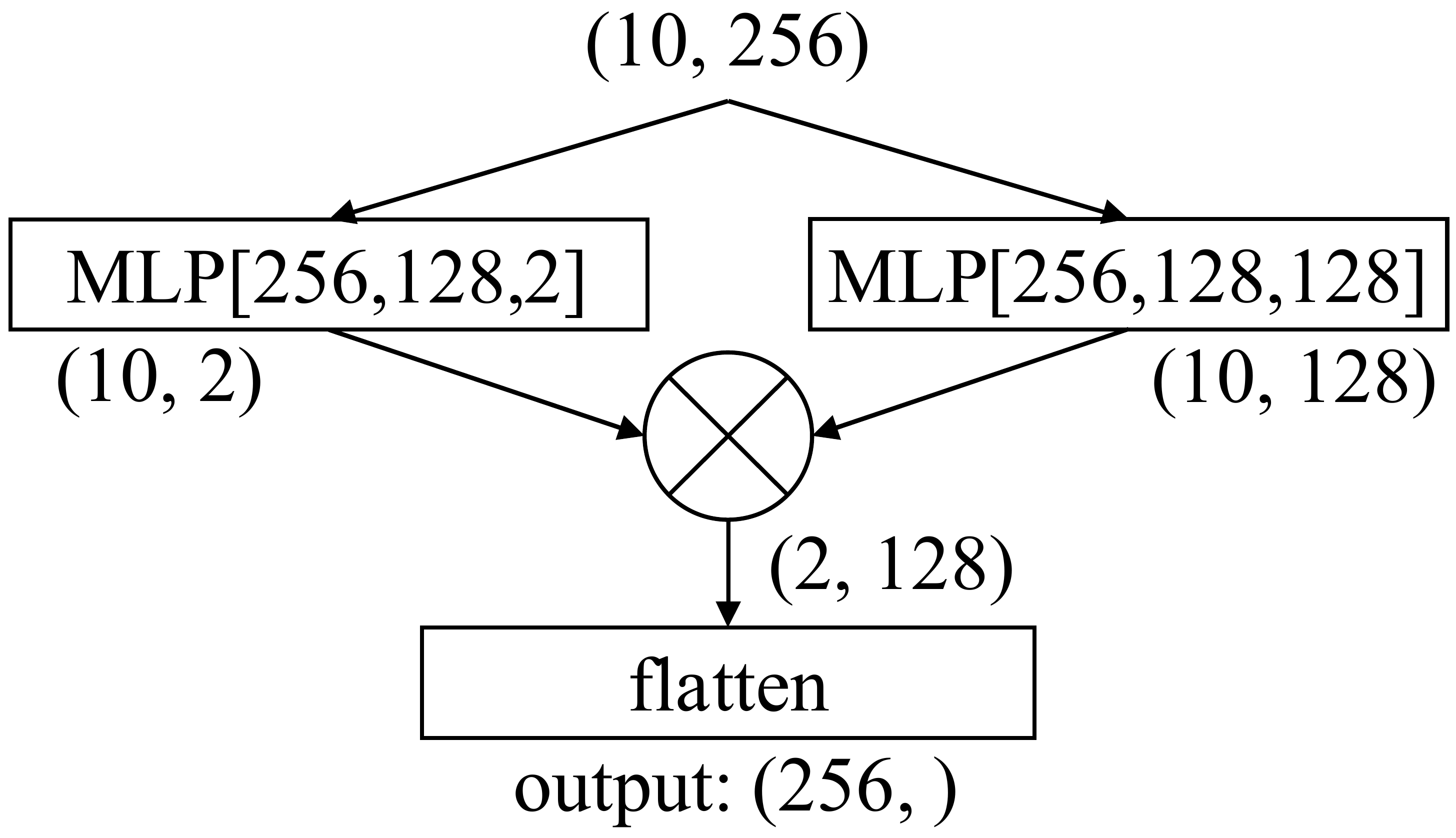}
  \centerline{(f) DuMLP-Pin}
  \caption{Detailed structure diagrams of all set feature extractors.}
  \label{fig:9}
\end{figure}

In some previous work, CelebA is also selected as the benchmark dataset \citep{zaheer2017deep,lee2019set}. Considering that they employ different sampled data and backbone methods to extract features from images, it seems difficult and unfair if making a comparison with them. For this reason, we resample our datasets. First, two attributes are selected randomly and 9 images with attributes and 1 image without any attribute are then chosen randomly. Each image appears at most once in our sampled dataset. Second, we resize all images from $218\times 178$ to $49\times 40$ to balance the number of parameters between the CNN backbone and the other part. The CNN backbone consists of 6 convolutional layers. Each convolutional layer is followed by batch normalization and ReLU. There are max-pooling layers with the kernel size of 2 between the 2nd and 3rd convolutional layers and the 4th and 5th convolutional ones, respectively. Finally, the average-pooling is exploited after the final convolutional layer. The total number of parameters is about 1.48 million. The details of all architectures are shown in Figure \ref{fig:9}.

\begin{figure}[htbp]
  \centering
  \includegraphics[width=0.43\linewidth]{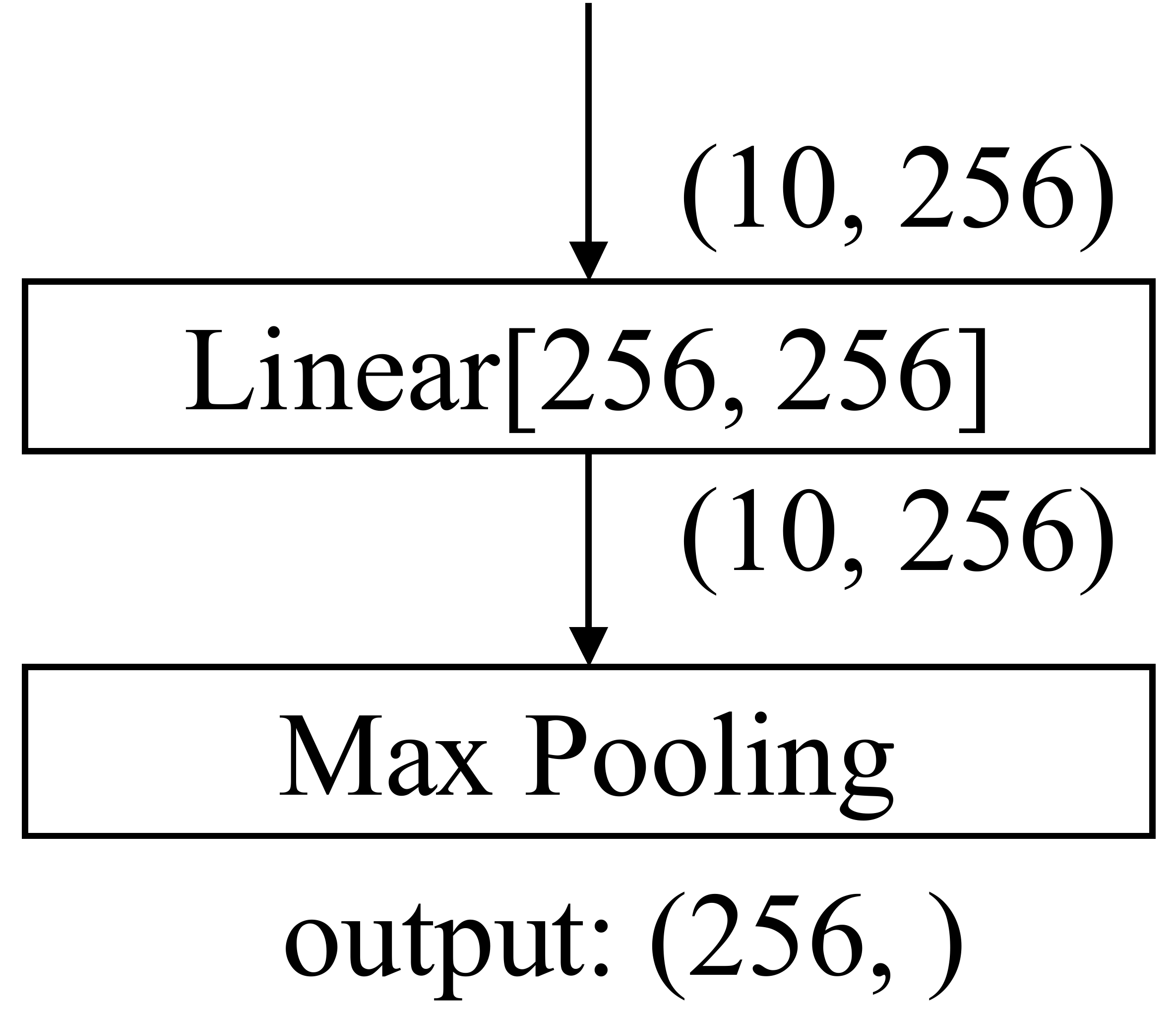}
  \centerline{(a) Max-pooling}
  \includegraphics[width=0.43\linewidth]{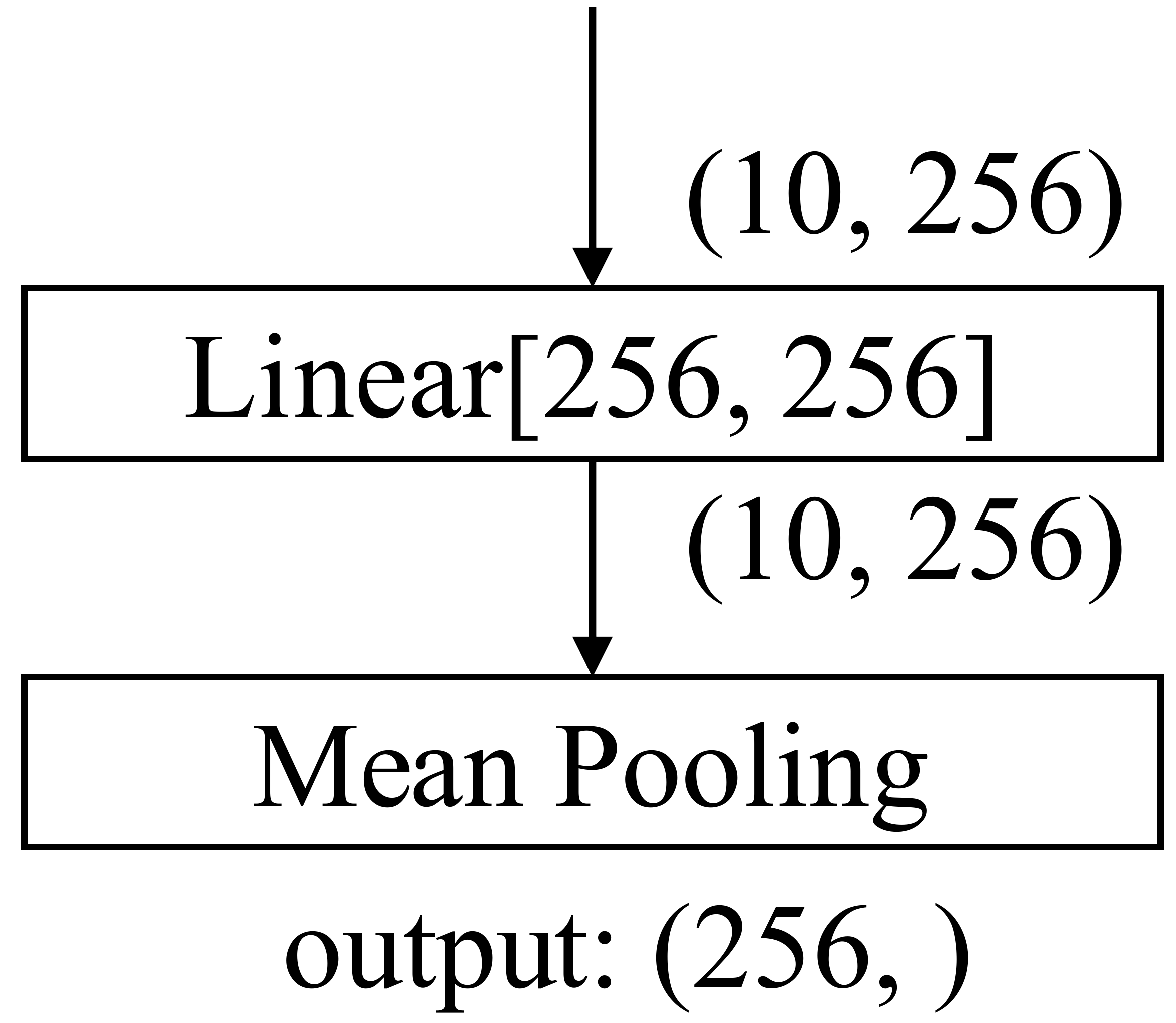}
  \centerline{(b) Mean-pooling}
  \includegraphics[width=0.43\linewidth]{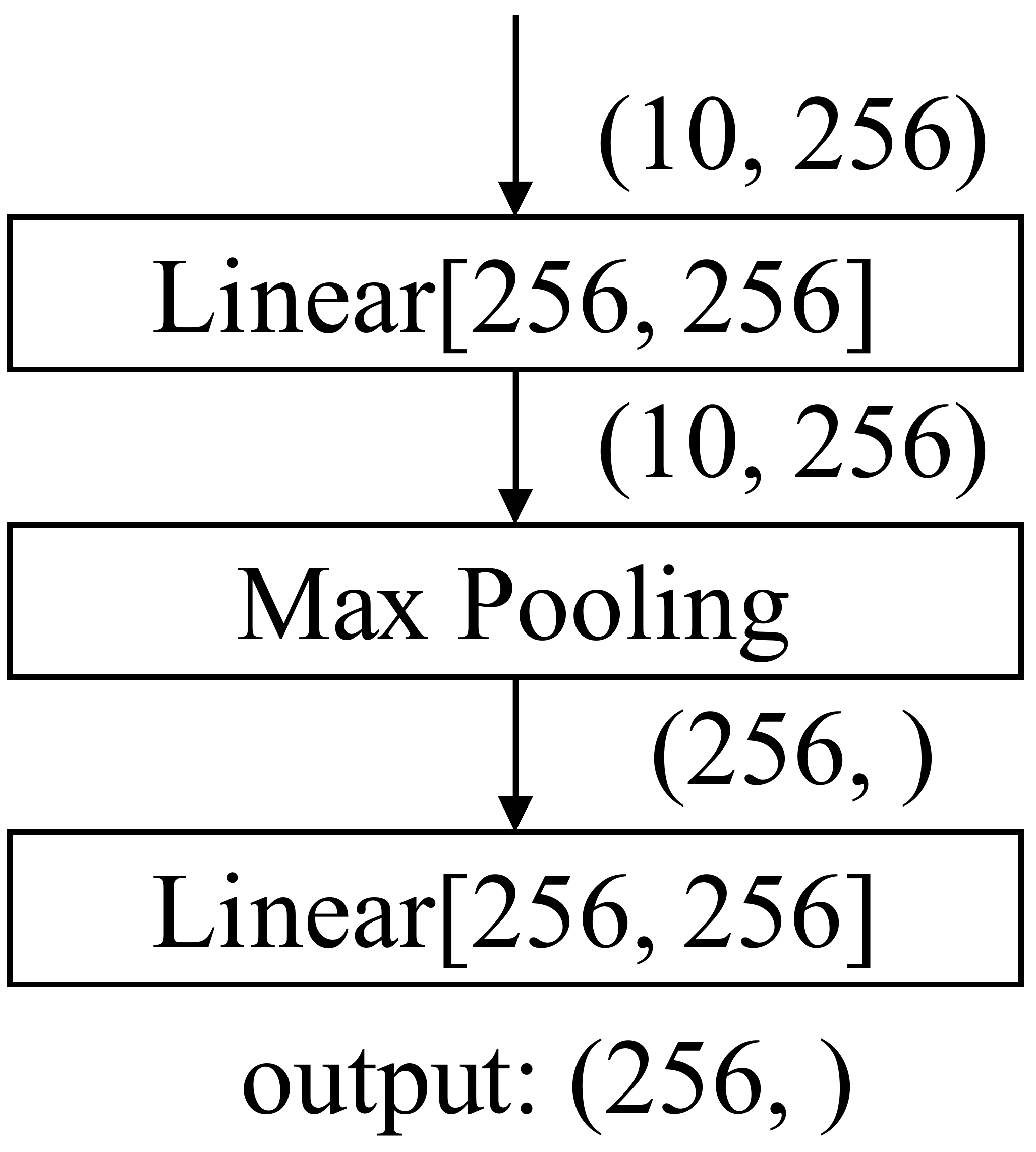}
  \centerline{(c) PointNet}
\end{figure}

We employ grid search for $h$ and $\gamma$ in Deep Sets and PointNet, which are done from identity mappings, linear mappings [256, 256], MLP [256, 64, 256], MLP [256, 128, 256], and MLP [256, 256, 256]. For Set Transformer, we make use of grid search for the number of heads from 1, 2, 4, 8, 16, 32, 64, 128, and 256. For DuMLP-Pin, we search for the factorization from $2\times 128$, $4\times 64$, and $8\times 32$. Note that $16\times 16$ is not feasible due to $10<16$. Since grid search is done with one random seed, average results may be different. Similarly, we adopt softmax as the activation function in one MLP, while there is no activation function in the other MLP. The batch size is selected as 8. We train all models for 30 epochs. The initial learning rate is set to 0.01 and is dropped by 10x at epoch 10.

\subsection{Point Cloud Classification}

\begin{figure}[htbp]
  \centering
  \includegraphics[width=0.75\linewidth]{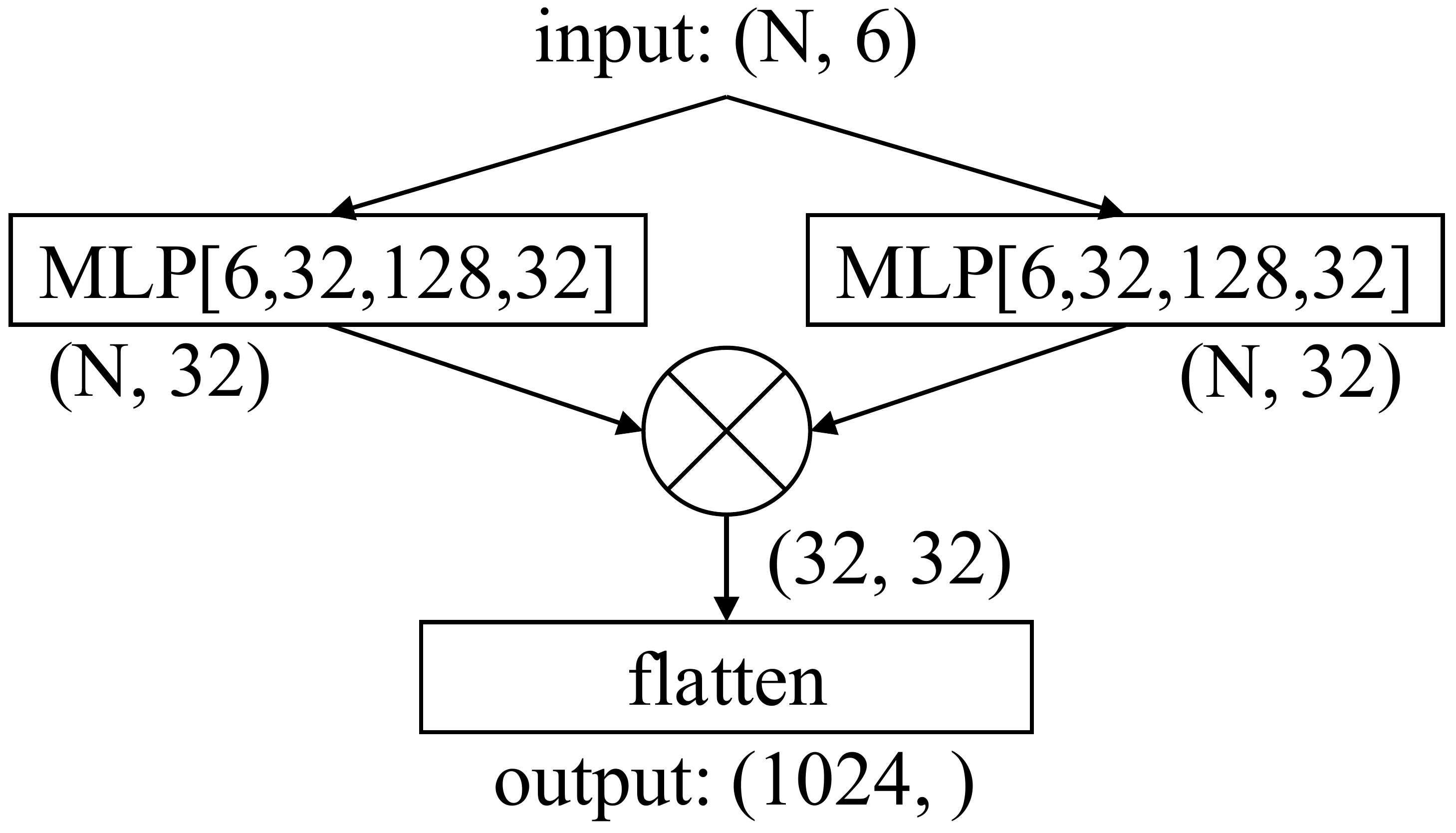}
  \centerline{(a) Aggregation block}
  \includegraphics[width=0.42\linewidth]{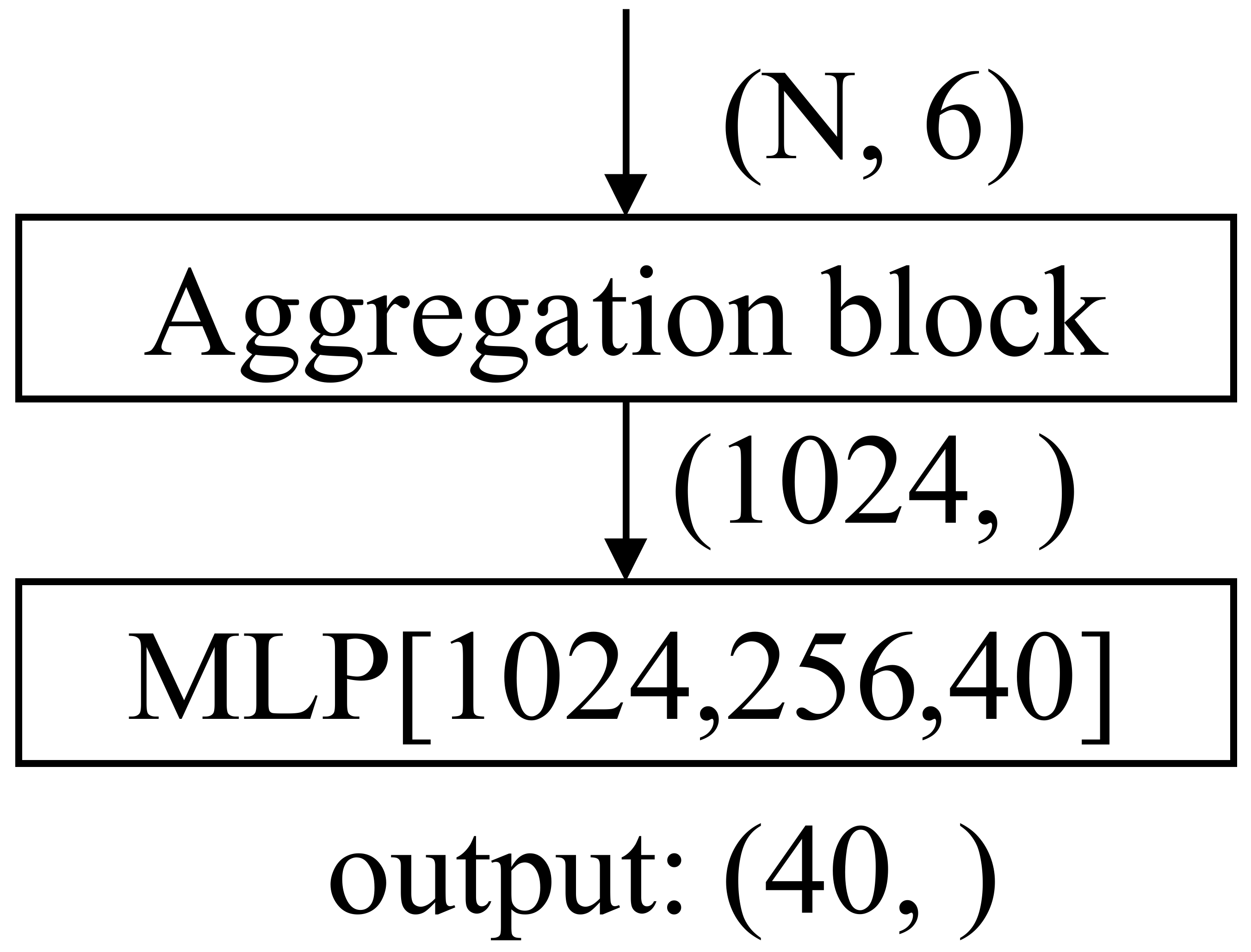}
  \centerline{(b) DuMLP-Pin}
  \caption{Detailed structure diagrams of the DuMLP-Pin applied in ModelNet40.}
  \label{fig:7}
\end{figure}

The structure diagram of DuMLP-Pin is shown in Figure \ref{fig:7}. The batch size is set to 32. We train the model for 250 epochs. The initial learning rate is set to 0.01 and is dropped by 10x at epoch 200. In one MLP, softmax is exploited as the activation function, while there is no activation function in another MLP. Note that all point cloud data is normalized before being fed into MLPs. In addition, we adopt several classic data augmentations, including random drop, random scale, random shift, random rotation, and additional Gaussian noise.

\subsection{Point Cloud Part Segmentation}

\begin{figure}[!t]
  \centering
  \includegraphics[width=0.75\linewidth]{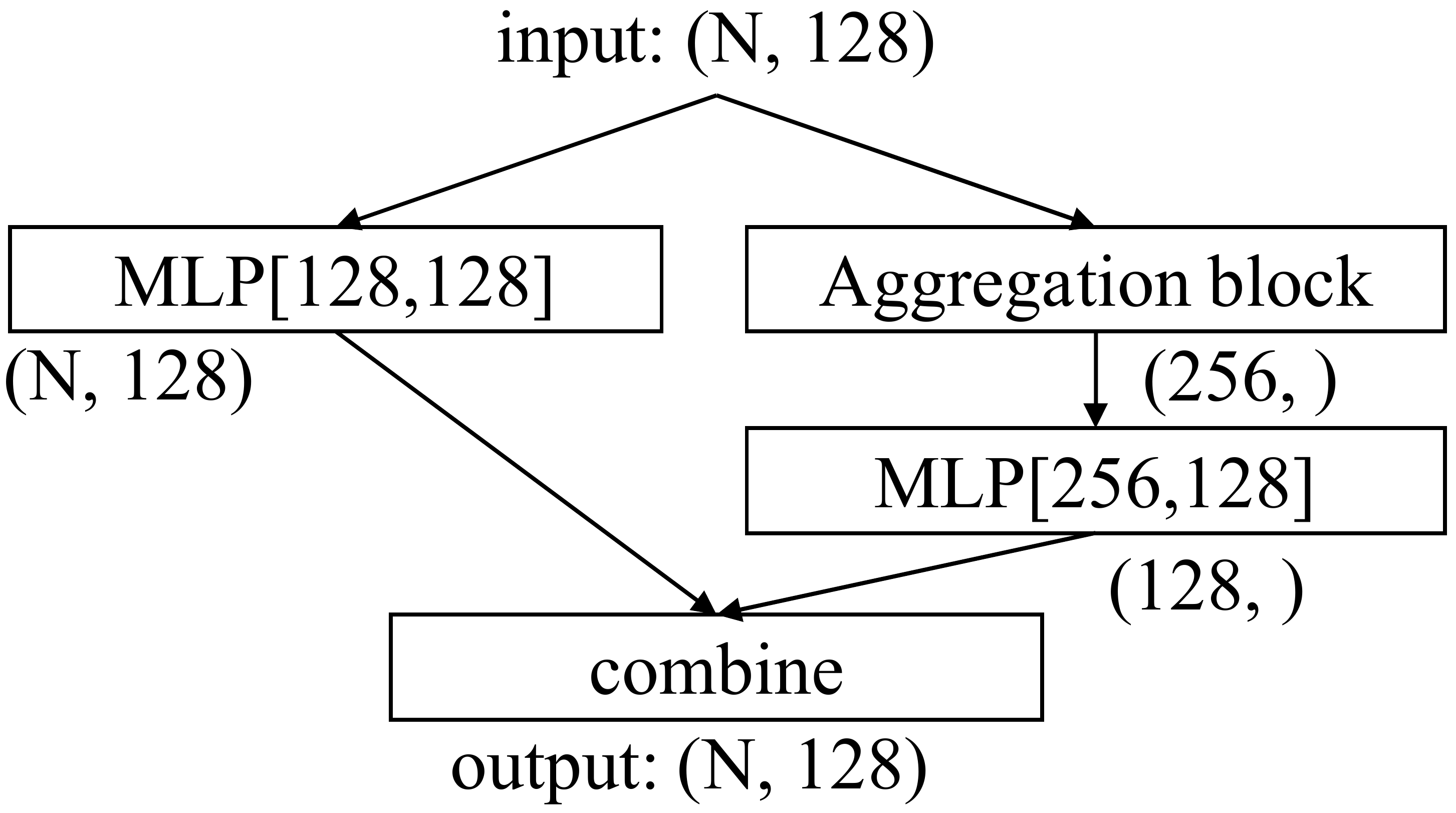}
  \centerline{(c) Broadcast block II}
  \includegraphics[width=0.45\linewidth]{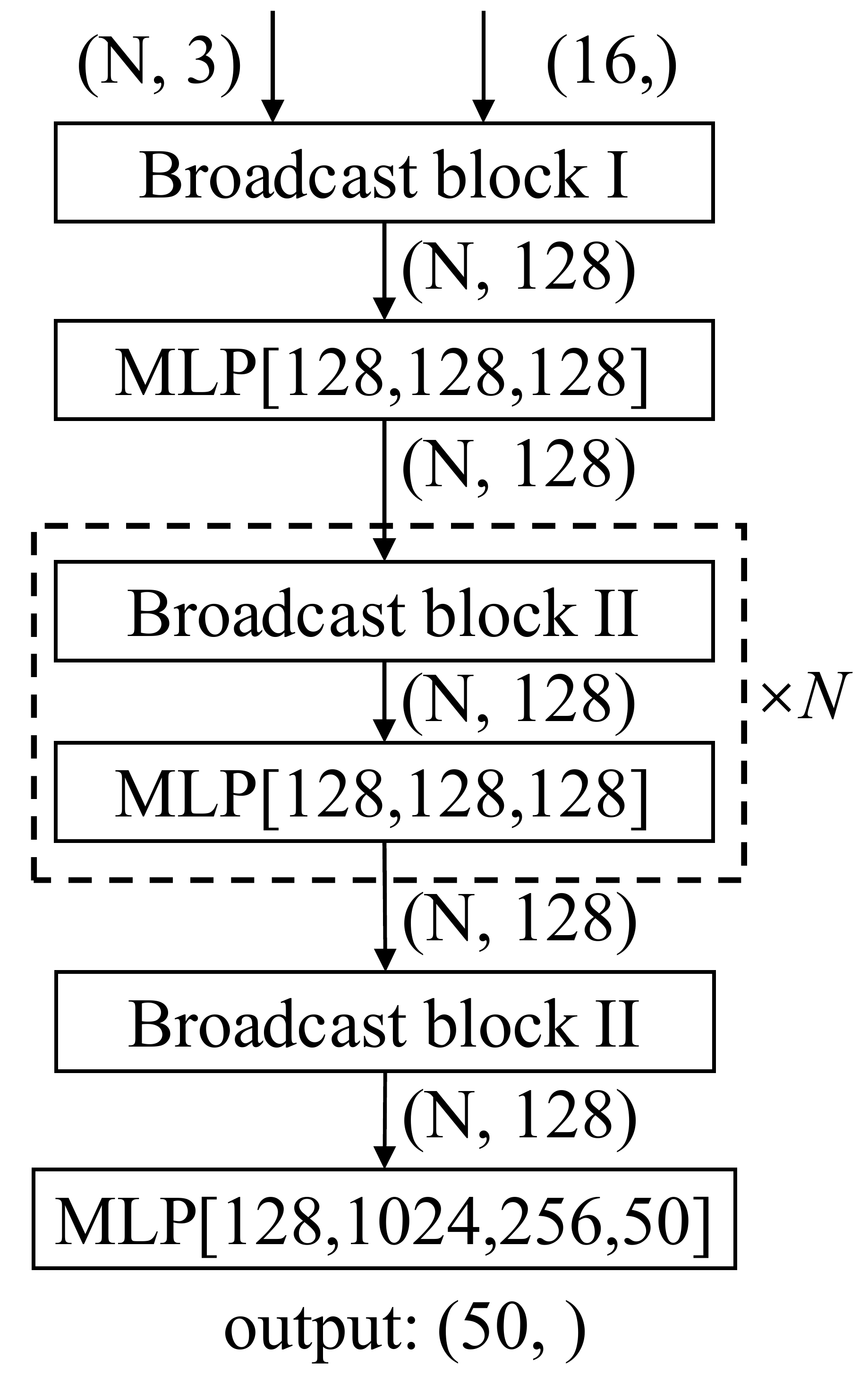}
  \centerline{(d) DuMLP-Pin-S ($N=0$)}
  \centerline{ DuMLP-Pin-L ($N=2$)}
  \caption{Detailed structure diagrams of the DuMLP-Pin applied in ShapeNetPart.}
  \label{fig:8}
\end{figure}

The structure diagram of DuMLP-Pin used in the ShapeNetPart experiment is shown in Figure \ref{fig:8}. The batch size is set to 32. We train the model for 400 epochs. The initial learning rate is set to 0.01 and is dropped by 10x at epoch 200. In one MLP, softmax is exploited as the activation function, while there is no activation function in another MLP. We adopt several classic data augmentations, including random scale and random shift.

\begin{figure}[!b]
  \centering
  \includegraphics[width=0.75\linewidth]{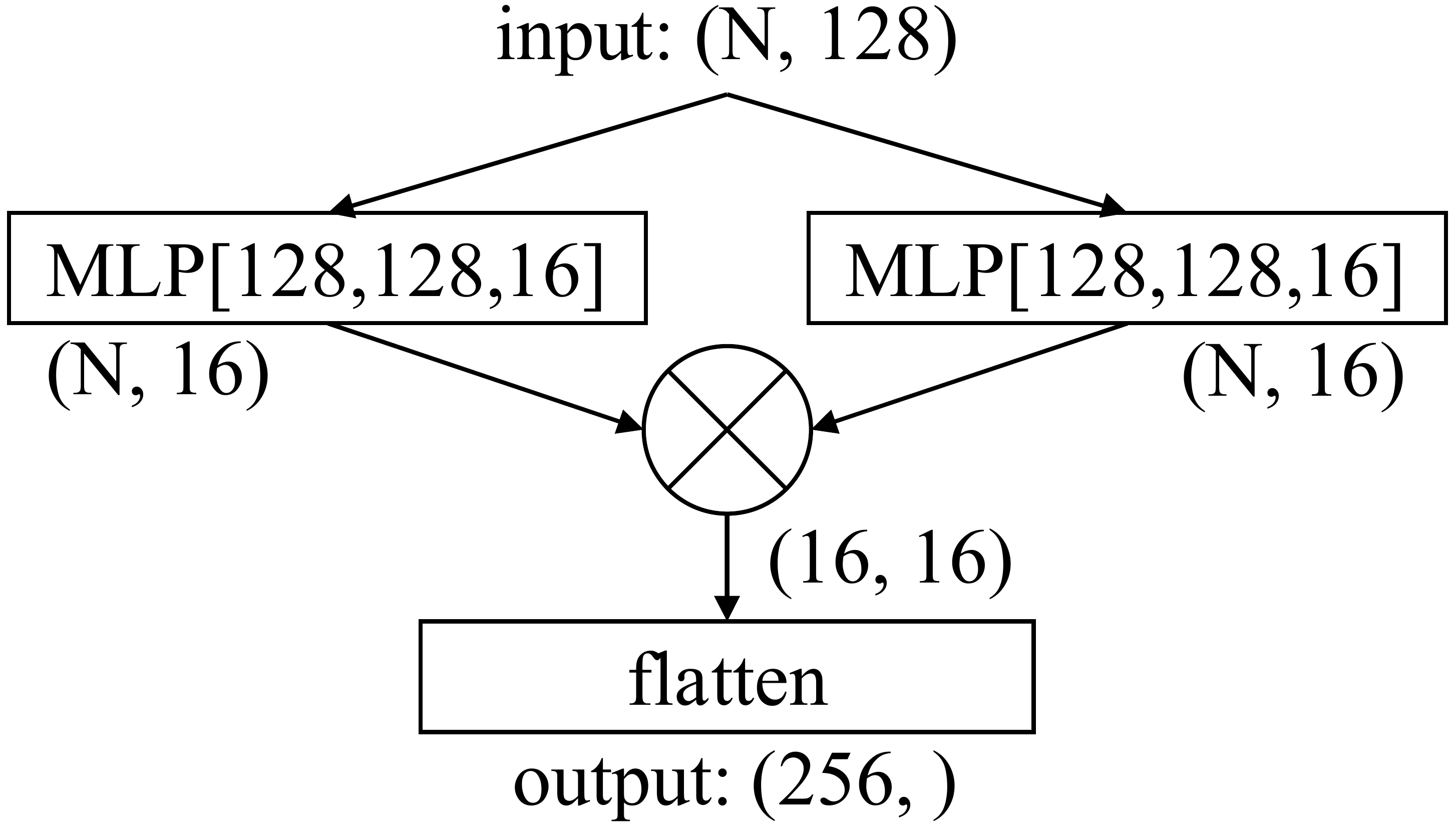}
  \centerline{(a) Aggregation block}
  \includegraphics[width=0.75\linewidth]{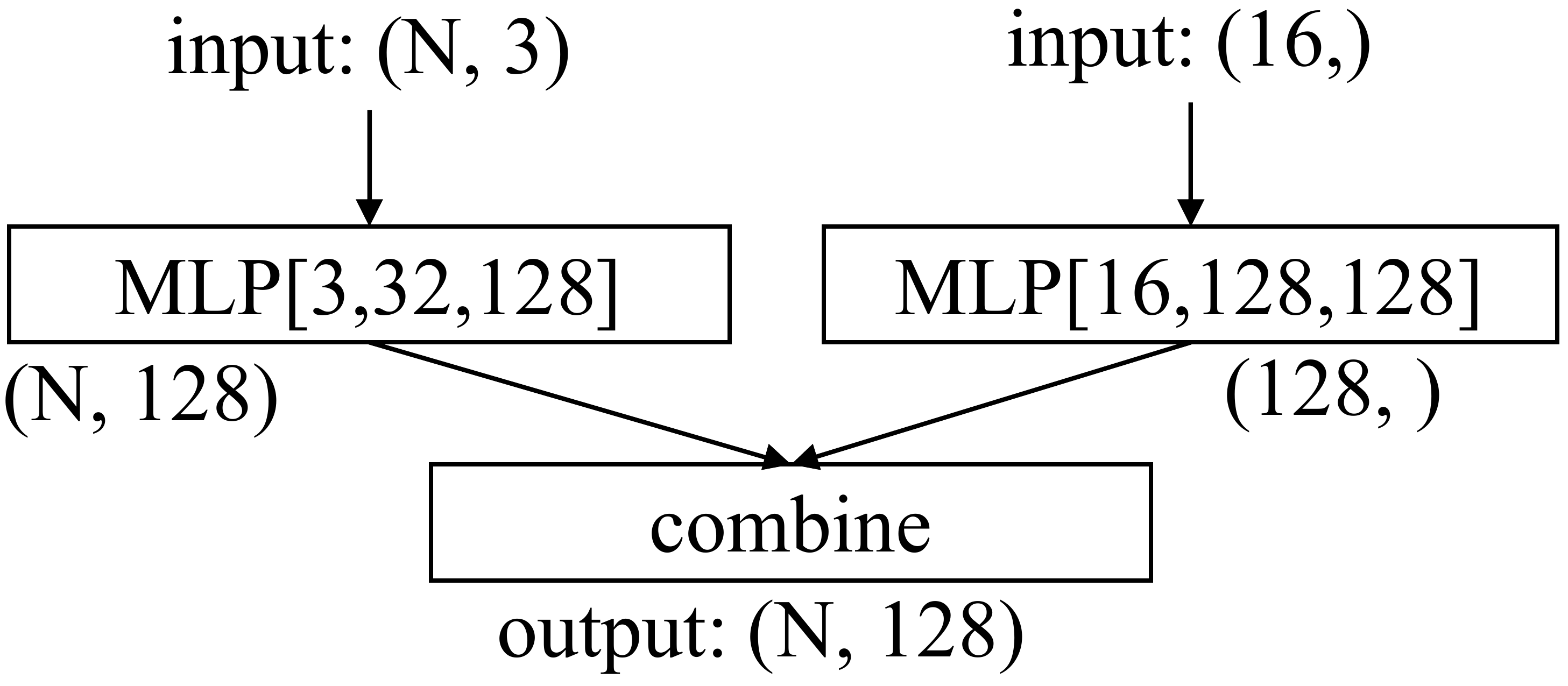}
  \centerline{(b) Broadcast block I}
\end{figure}

Besides DuMLP-Pin-S and DuMLP-Pin-L in the ShapeNetPart experiment, we complete experiments by increasing the number of broadcast blocks from 2 to 7. It is easy to see from Table \ref{tab:9} that the best performance could be reached for 4 broadcast blocks. In fact, we may find different optimal numbers for different tasks. When the number of broadcast blocks is too small, the representation capability is limited. On the contrary, it is difficult to train a model with competitive results.

\begin{table}[htbp]
  \centering
  \begin{tabular}{cc}
    \toprule
    Number of blocks  & mIoU(\%) \\
    \midrule
    2 & 83.5 \\
    3 & 84.7 \\
    4 & \textbf{84.9} \\
    5 & 84.6 \\
    6 & 84.2 \\
    7 & 83.2 \\
    \bottomrule
  \end{tabular}
  \caption{Influence of number of broadcast blocks.}
  \label{tab:9}
\end{table}

\subsection{Point Cloud Semantic Segmentation}

Besides part segmentation, we also evaluate DuMLP-Pin in the point cloud semantic segmentation task using the ScanNet. The ScanNet \cite{dai2017scannet} is an indoor scene dataset with 1,201 scenes in the training set and 312 scenes in the test set. We use the same sampled and preprocessed data in \citeauthor{qi2017pointnet++}, which means that RGB information is removed. We take the per-voxel accuracy as the evaluation metric. All results quoted are taken from the cited papers. Still, two models of different sizes are evaluated. From Table \ref{tab:8}, DuMLP-Pin-S performs better than PointNet while both DuMLP-Pin-S and DuMLP-Pin-L are significantly worse than PointNet++. For a large scene, global aggregation methods consider all points, making it hard or even impossible to divide local points into meaningful parts. Therefore, DuMLP-Pin outperforms other global aggregation methods like PointNet but still underperforms local aggregation methods.

\begin{table}[htb]
  \centering
  \begin{tabular}{ccc}
    \toprule
    Method & \#Params(M) & mIoU(\%) \\
    \midrule
    PointNet \shortcite{qi2017pointnet} & 3.54 & 73.0\\
    PointNet++ \shortcite{qi2017pointnet++} & 0.85 & 84.5\\
    \midrule
    DuMLP-Pin-S & 0.71 & 76.9 $\pm$ 0.04 \\
    DuMLP-Pin-L & 0.83 & 77.6 $\pm$ 0.06 \\
    \bottomrule
  \end{tabular}
  \caption{The ScanNet point cloud semantic segmentation results.}
  \label{tab:8}
\end{table}

\begin{figure}[!b]
  \centering
  \includegraphics[width=0.75\linewidth]{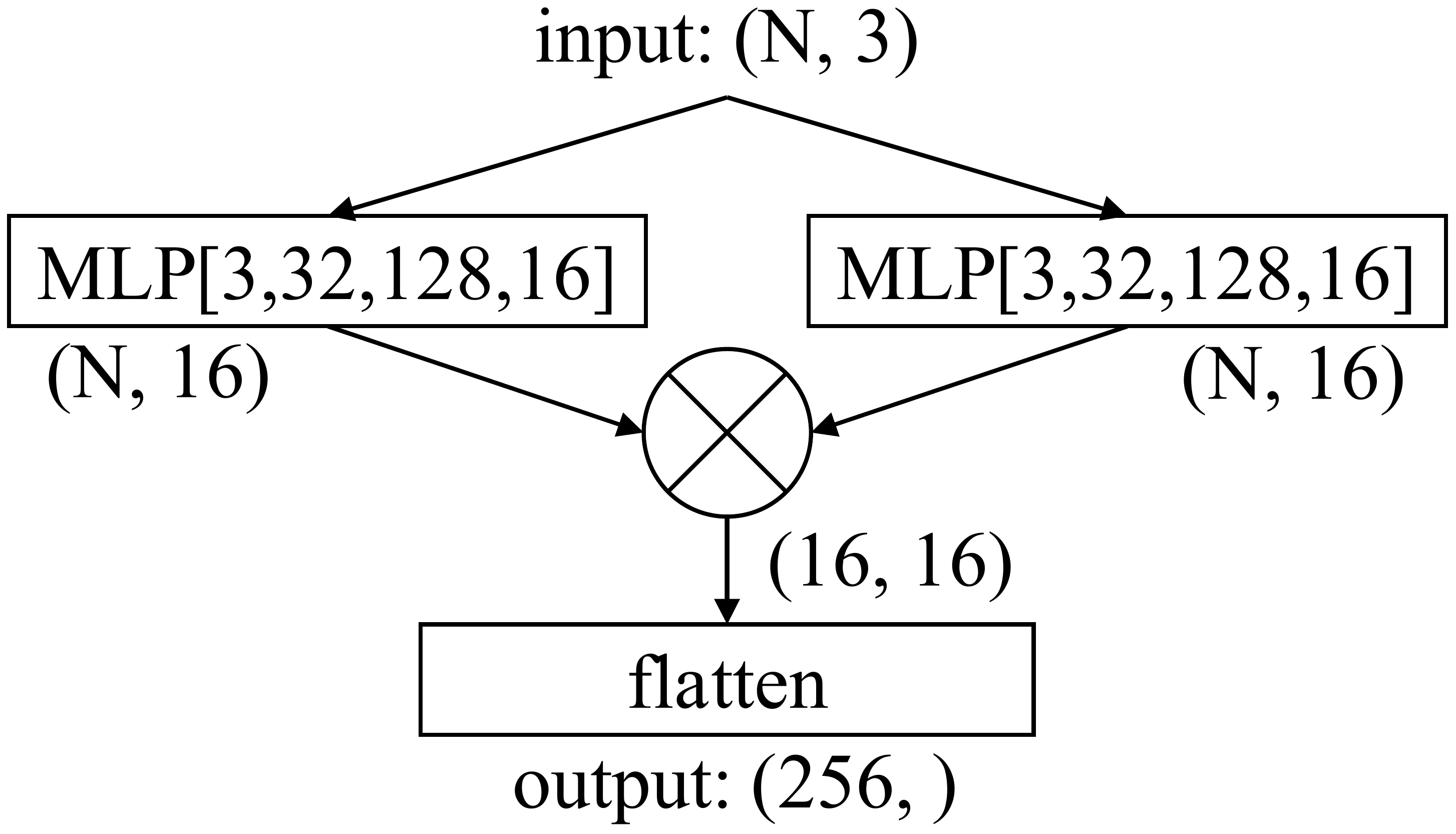}
  \centerline{(a) Aggregation block I}
  \includegraphics[width=0.75\linewidth]{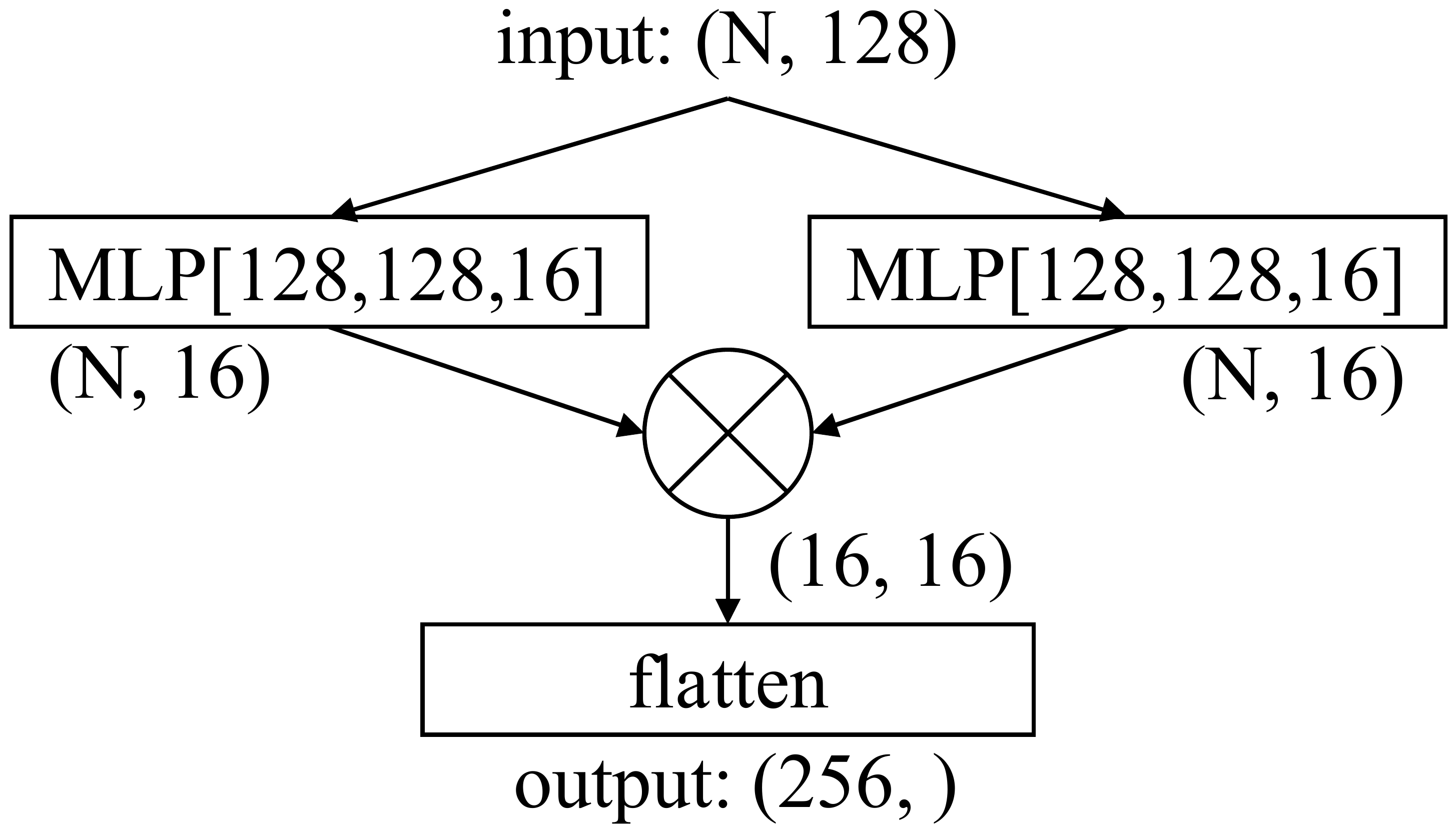}
  \centerline{(b) Aggregation block II}
  \includegraphics[width=0.75\linewidth]{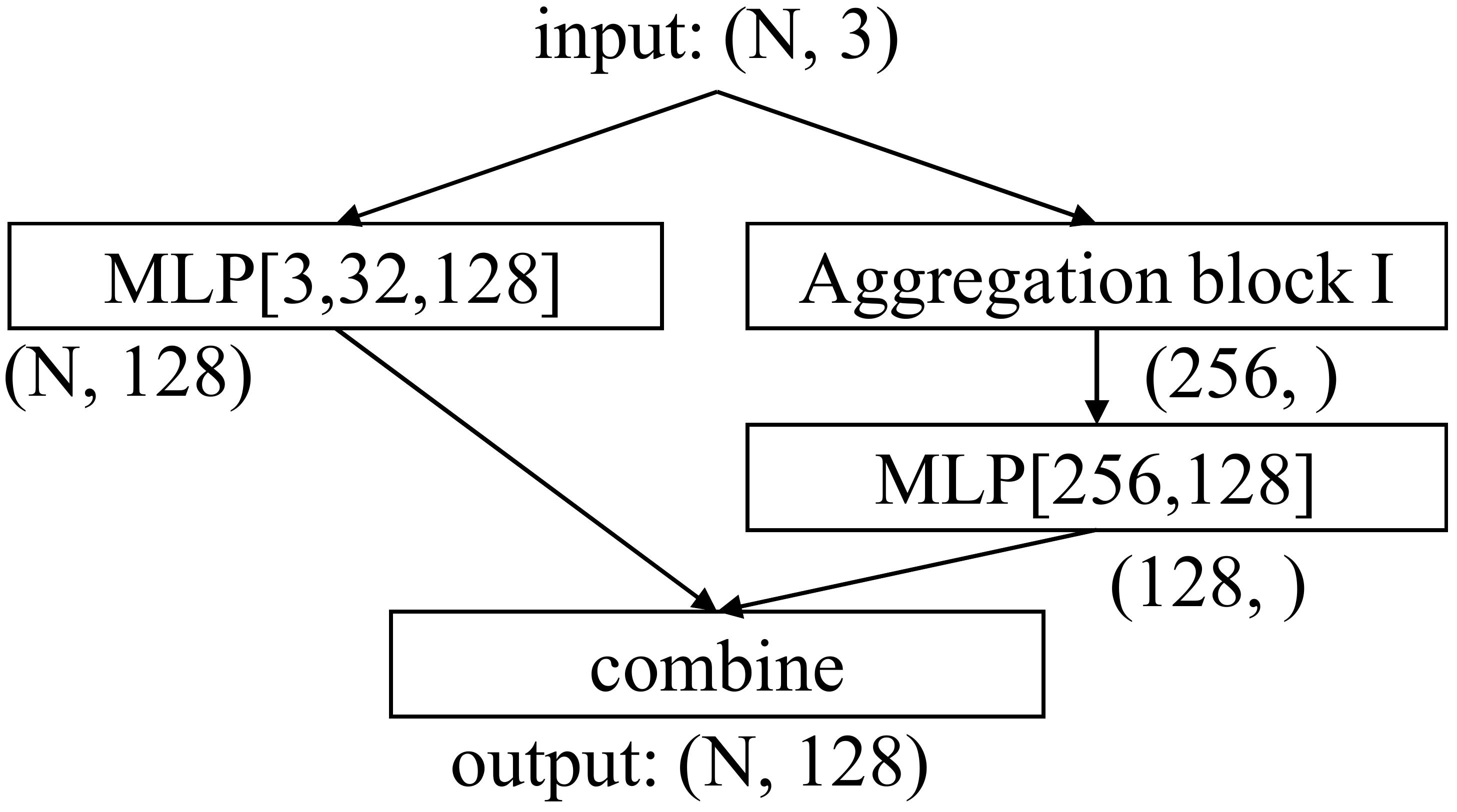}
  \centerline{(c) Broadcast block I}
\end{figure}

\begin{figure}[!t]
  \centering
  \includegraphics[width=0.75\linewidth]{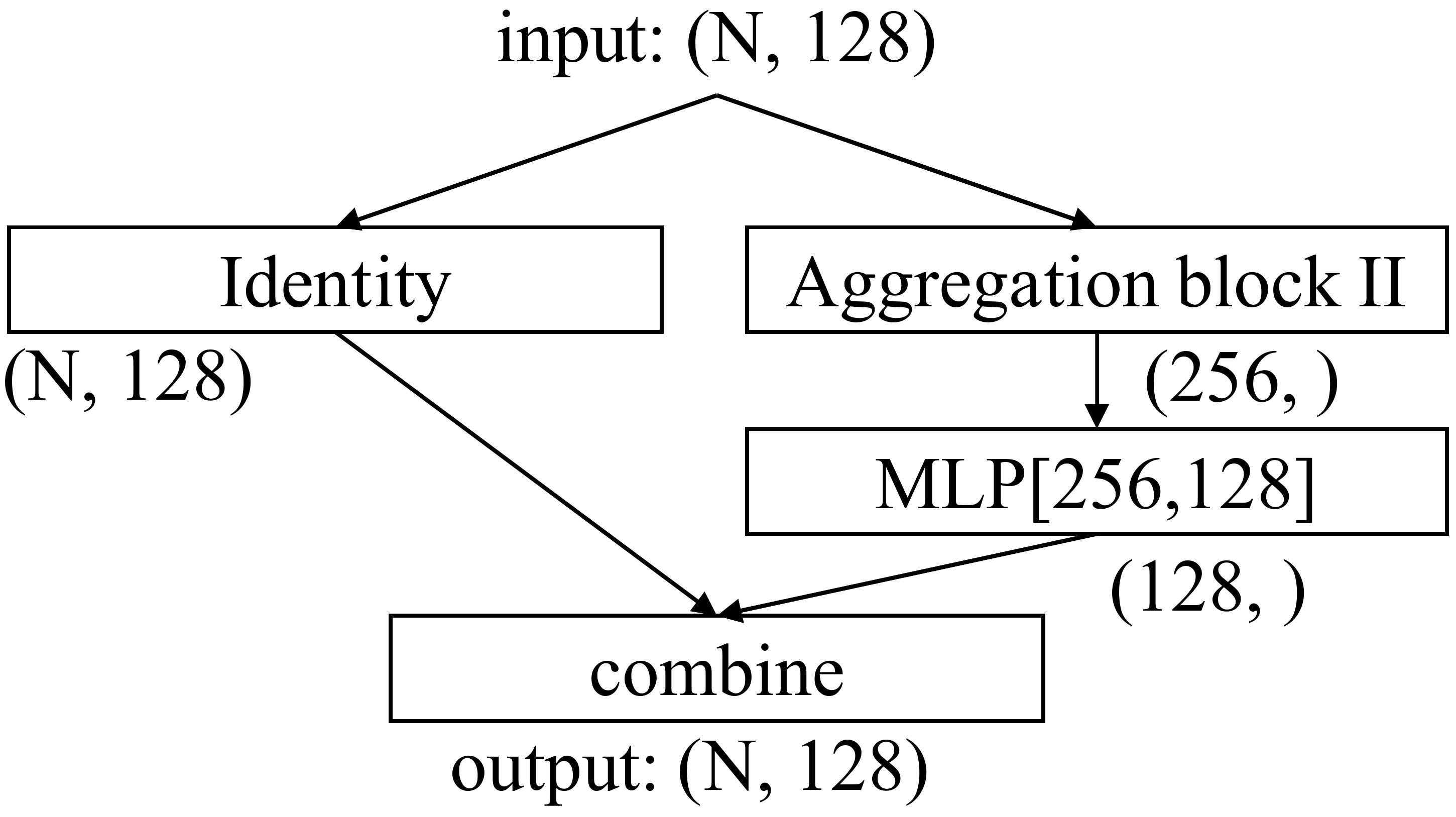}
  \centerline{(d) Broadcast block II}
  \includegraphics[width=0.45\linewidth]{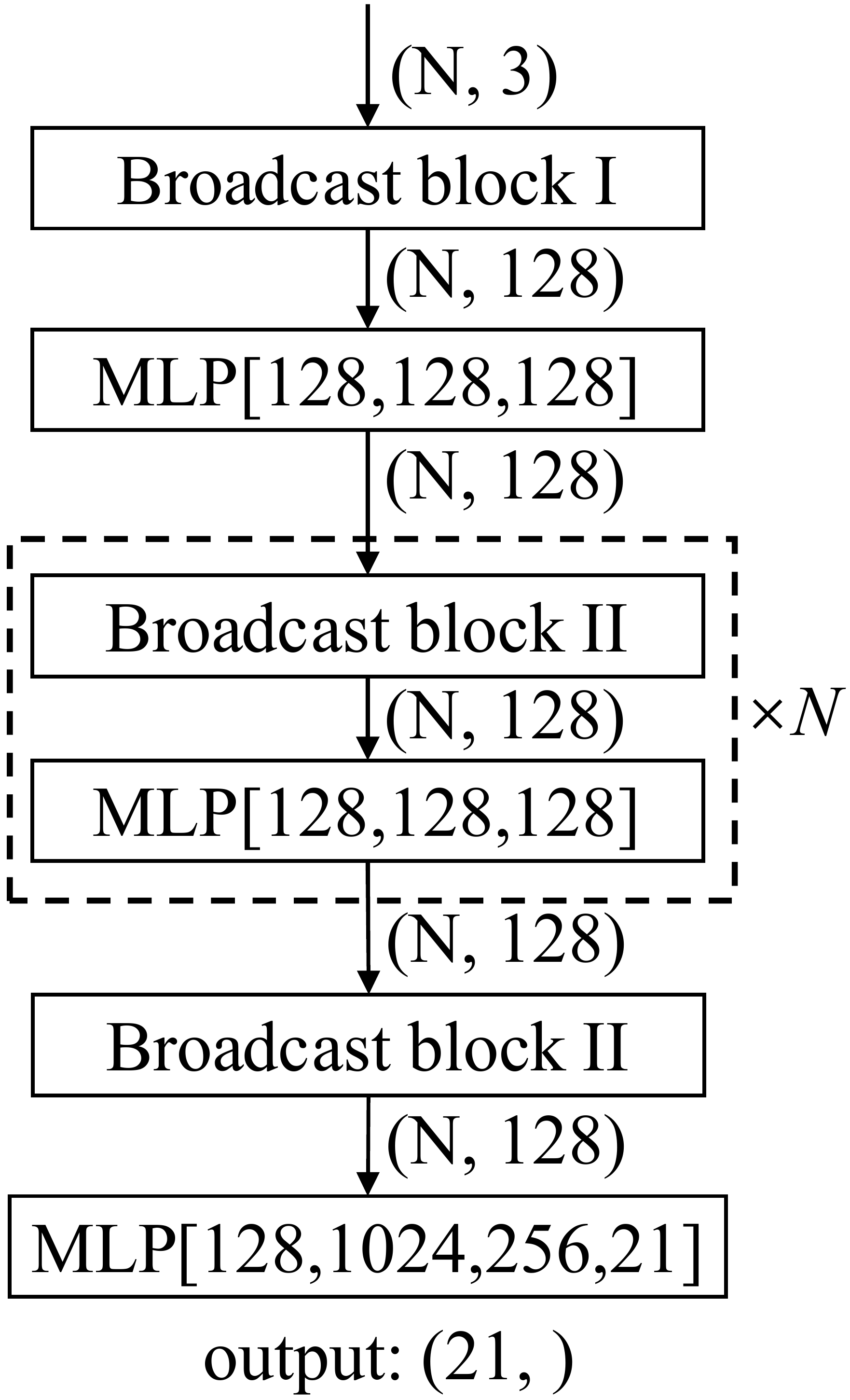}
  \centerline{(e) DuMLP-Pin-S ($N=0$)}
  \centerline{DuMLP-Pin-L ($N=1$)}
  \caption{Detailed structure diagrams of the DuMLP-Pin applied in ScanNet.}
  \label{fig:10}
\end{figure}

The structure diagram of DuMLP-Pin used in the ScanNet experiment is shown in Figure \ref{fig:10}. The batch size is set to 32. We train the model for 500 epochs. The initial learning rate is set to 0.0001 and it is increased to 0.01 by 20 epochs. The learning rate is dropped by 10x at epoch 220. In one MLP, softmax is exploited as the activation function, while there is no activation function in another MLP. We adopt random drop as the data augmentation method.

The main difference between the ShapeNetPart and the ScanNet is that the former is a small-scale object dataset and the latter is a large-scale scene dataset. Global aggregation methods totally omit the local structures, which is more important in semantic scene segmentation than in object part segmentation. Therefore, as the scale grows, the result from the corresponding experiment dampens. Even segmentation tasks are more suitable for local aggregation methods, DuMLP-Pin still shows its potential in the object-level segmentation task.

\bibliography{aaai22}